\documentclass[preprint,12pt]{elsarticle}

\usepackage[T1]{fontenc} 
\usepackage{amssymb}
\usepackage{amsmath}
\usepackage{bm}
\usepackage{algorithm}
\usepackage{algorithmic}
\usepackage{multirow}
\usepackage{subcaption} 
\usepackage[colorlinks,
            linkcolor=blue,      
            anchorcolor=blue,  
            citecolor=blue,       
            ]{hyperref}

\journal{Nuclear Physics B}

\begin{document}

\begin{frontmatter}



\title{Graph Neural Networks with Coarse- and Fine-Grained Division for Mitigating Label Sparsity and Noise} 

\author{Shuangjie Li}\ead{shuangjieli@smail.nju.edu.cn}
\author{Baoming Zhang}
\author{Jianqing Song}
\author{Gaoli Ruan}
\author{\corref{cor1}Chongjun Wang}\ead{chjwang@nju.edu.cn}
\author{Junyuan Xie}


\begin{abstract}
Graph Neural Networks (GNNs) have gained considerable prominence in semi-supervised learning tasks in processing graph-structured data, primarily owing to their message-passing mechanism, which largely relies on the availability of clean labels.
However, in real-world scenarios, labels on nodes of graphs are inevitably noisy and sparsely labeled, significantly degrading the performance of GNNs.
Exploring robust GNNs for semi-supervised node classification in the presence of noisy and sparse labels remains a critical challenge. 
Therefore, we propose a novel \textbf{G}raph \textbf{N}eural \textbf{N}etwork with \textbf{C}oarse- and \textbf{F}ine-\textbf{G}rained \textbf{D}ivision for mitigating label sparsity and noise, namely GNN-CFGD.
The key idea of GNN-CFGD is reducing the negative impact of noisy labels via coarse- and fine-grained division, along with graph reconstruction.
Specifically, we first investigate the effectiveness of linking unlabeled nodes to cleanly labeled nodes, demonstrating that this approach is more effective in combating labeling noise than linking to potentially noisy labeled nodes.
Based on this observation, we introduce a Gaussian Mixture Model (GMM) based on the memory effect to perform a coarse-grained division of the given labels into clean and noisy labels. 
Next, we propose a clean labels oriented link that connects unlabeled nodes to cleanly labeled nodes, aimed at mitigating label sparsity and promoting supervision propagation.
Furthermore, to provide refined supervision for noisy labeled nodes and additional supervision for unlabeled nodes, we fine-grain the noisy labeled and unlabeled nodes into two candidate sets based on confidence, respectively.
Extensive experiments on various datasets demonstrate the superior effectiveness and robustness of GNN-CFGD.
\end{abstract}

\begin{keyword}
Graph Neural Network \sep Semi-Supervised Node Classification \sep Label Noise



\end{keyword}

\end{frontmatter}


\section{Introduction}
\label{sec1}
Graph-structured data is ubiquitous in various real-world scenarios, such as social media \citep{krishnamurthy2008few, zhang2019your}, recommendation systems \citep{wu2022graph}, and power networks \citep{holmgren2006using}.
In recent years, Graph Neural Networks (GNNs) have gained considerable prominence due to their effectiveness in processing graph-structured data \citep{kipf2016semi, velivckovic2017graph, wu2020comprehensive, wu2022graph} by utilizing a message-passing mechanism \citep{gilmer2017neural}, where node embeddings are derived by aggregating and transforming the embeddings of neighboring nodes. This mechanism facilitates information propagation along the graph structure, benefiting downstream tasks such as semi-supervised node classification \citep{wang2020gcn, kipf2016semi}.

Despite their promising performance, most existing GNNs assumed that the observed labels are clean \citep{xu2018representation, wu2019simplifying, chen2024learning}.
However, in real-world scenarios, labels on nodes of graphs are inevitably sparsely and noisily labeled \citep{qian2023robust, chen2023erase}.
For instance, in social media graphs, it is common for only a small percentage of users to contribute to label generation. 
Consequently, some labels may intentionally or unintentionally misrepresent the truth. Similarly, in crowd-sourced labeling tasks, such as annotating fake news or medical knowledge graphs, the annotation process can be labor-intensive and costly, leading to inevitable label errors due to subjective judgment. These challenges result in graphs with sparse and noisy node labels, which significantly degrades the performance of GNNs.
Therefore, developing robust GNNs for mitigating label sparsity and noise is an important and challenging problem.
\begin{figure}[htbp]
    \centering
    \includegraphics[width=0.5\textwidth]{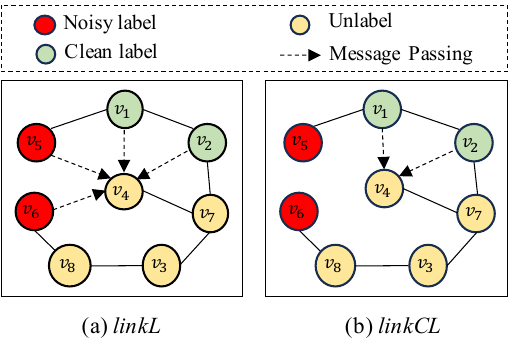}
    \caption{The illustration of message-passing for unlabeled node. (a) \textit{linkL} (such as NRGNN \citep{dai2021nrgnn} and RTGNN \citep{qian2023robust}): Directly linking unlabeled nodes to labeled ones may inadvertently connect to noisy labeled nodes; (b) \textit{linkCL}: Linking unlabeled nodes only to clean labeled nodes. We use unlabeled nodes $v_4$ as an example.}
    \label{fig_motivate1}
\end{figure}

\begin{figure}[htbp]
    \centering
    \begin{minipage}[b]{0.4\linewidth}
        \centering
        \includegraphics[width=\textwidth]{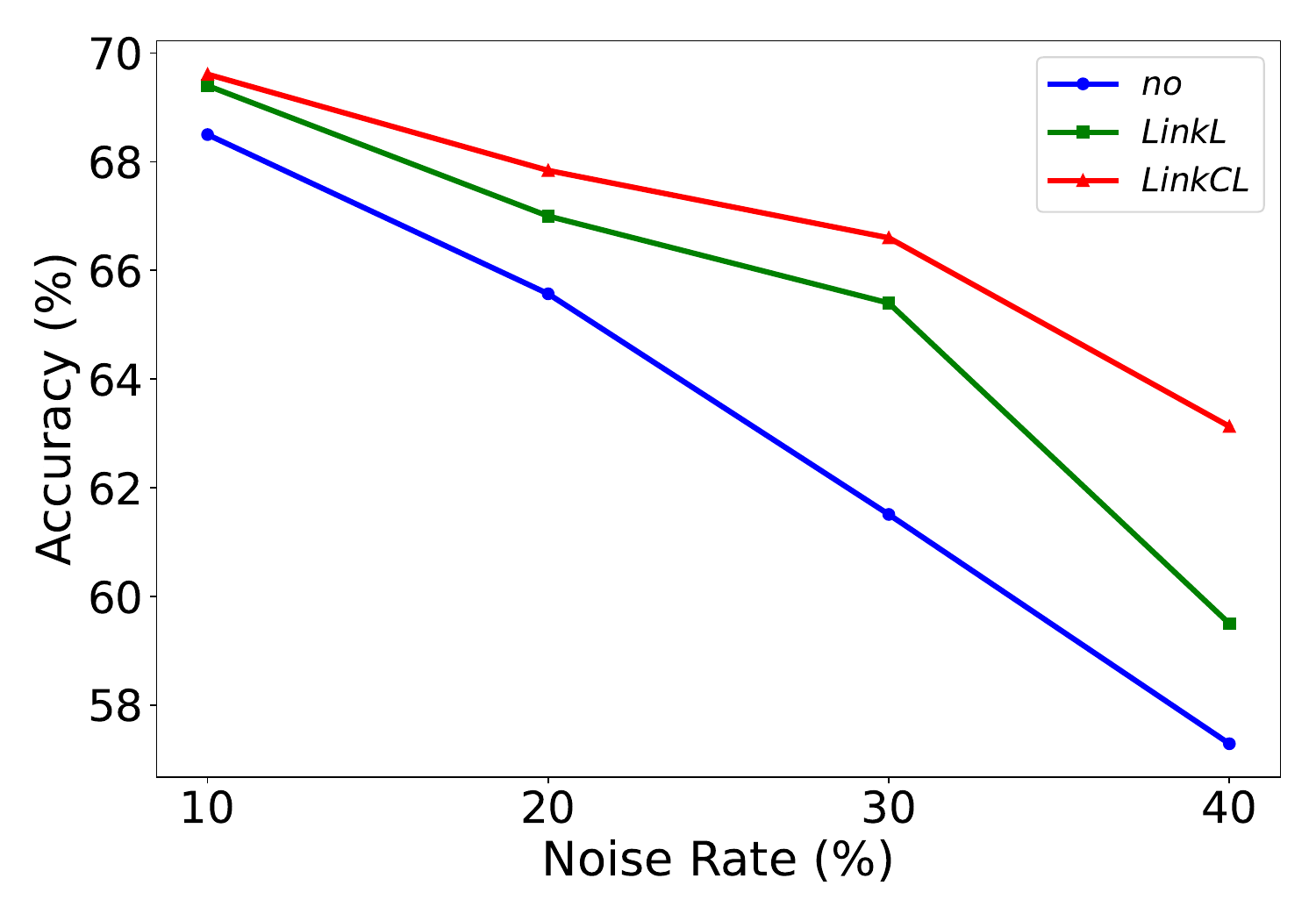}
        \centerline{\small{(a) Uniform noise}}
    \end{minipage}
    \begin{minipage}[b]{0.4\linewidth}
        \centering
        \includegraphics[width=\textwidth]{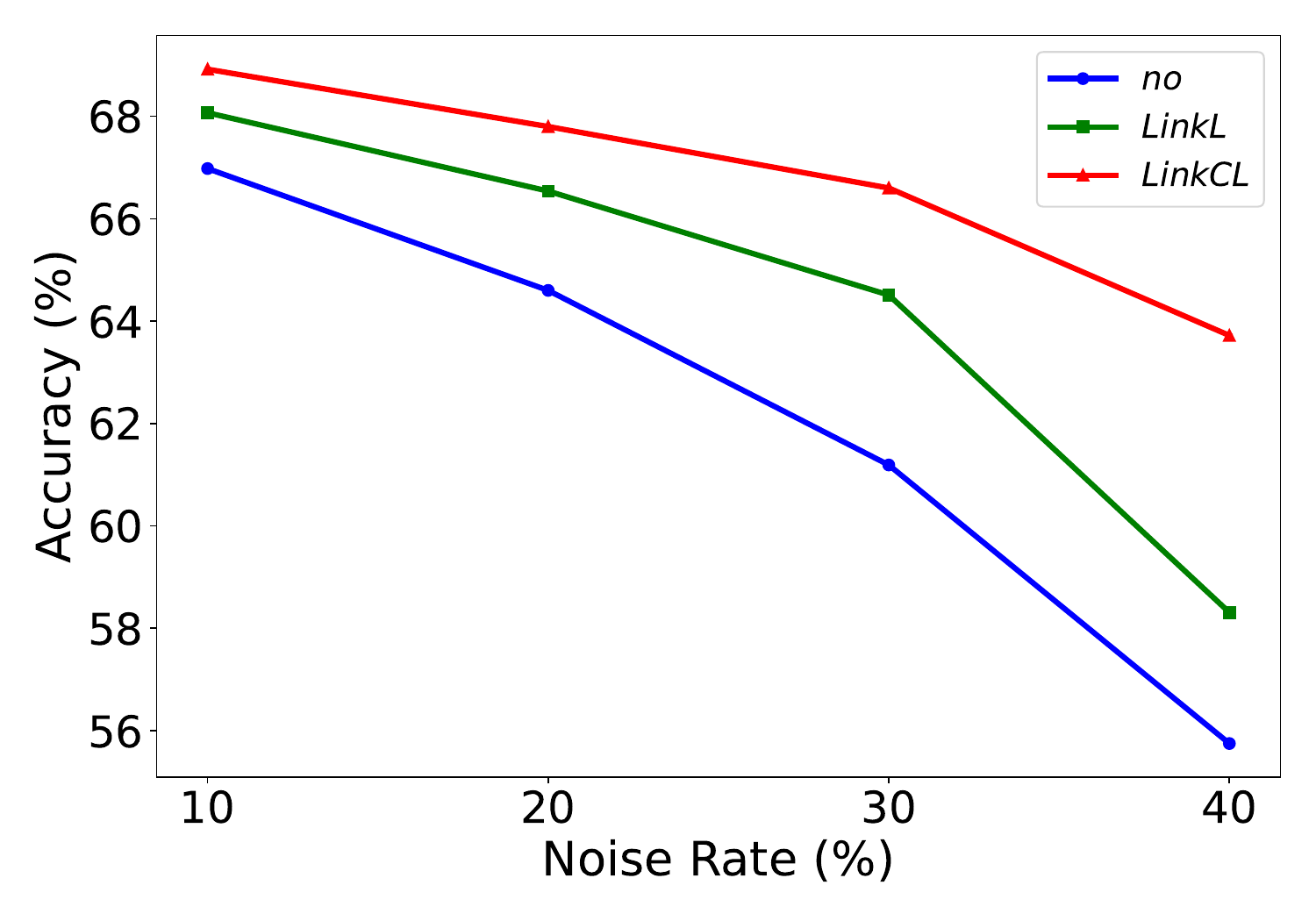}
        \centerline{\small{(b) Pair noise}}
    \end{minipage}
    \caption{The performance of GCN in node classification with different label noise rates on the Citeseer dataset with the setting: 1) for \textit{linkL}, we connect each unlabeled node to the top 50 labeled nodes that are most similar to it based on their features, and 2) for \textit{linkCL}, we connect each unlabeled node only to the clean nodes of top 50 labeled nodes that are most similar to it based on their features.}
    \label{fig_motivate2}
\end{figure}

To combat label noise, numerous methods have been proposed in the domain of Computer Vision (CV) \citep{li2020dividemix, yu2019does, han2018co}. However, these methods cannot be directly applied to address noisy labels on graphs. This limitation arises from the inherent complexity of graphs, where noisy information can propagate across the topology during the message-passing process. To mitigate label noise in graphs, D-GNN \citep{hoang2019learning} first proposed to train the neural network by optimizing the surrogate backward loss based on the correction matrix. However, this approach requires additional parameters to estimate the correction matrix effectively.
Subsequently, NRGNN \citep{dai2021nrgnn} was introduced to develop a label noise-resistant GNN by connecting unlabeled nodes to labeled ones. Furthermore, pseudo labels are utilized to reduce the sparse label issue. 
Recently, building on this idea, RTGNN \citep{qian2023robust} was proposed to facilitate robust graph learning by rectifying noisy labels and further generating pseudo-labels on unlabeled nodes.
However, it is worth noting that both methods share a common downside: directly linking unlabeled nodes to labeled ones may inadvertently connect to noisy labeled nodes, which exacerbates the propagation of noise to the unlabeled nodes along the added edges, as shown in Fig. \ref{fig_motivate1}.
To investigate the effect of different linking strategies on the node classification performance of GNNs, we design experiments with two types of noise (i.e., pair and uniform, see section \ref{subsubsec_dataset} for details) on the Citeseer dataset, as shown in Fig. \ref{fig_motivate2}. For each noise type, we select GCN \citep{kipf2016semi} as the backbone.

It is evident that for both types of label noise, as the noise ratio increases, the performance of GCN with \textit{linkL} and \textit{linkCL} outperforms that of GCN without added edges. Particularly noteworthy: the performance of GCN with \textit{linkCL} consistently remains the champion. Our experimental results also indicate that the proposed method reduces the number of noisily labeled nodes in \textit{linkCL}, as shown in \ref{visualization}. 
This leads to the following intuitive conclusion: 1) Connections between unlabeled nodes and noisily labeled nodes can result in the propagation of incorrect information; 2) Linking unlabeled nodes to cleanly labeled nodes is more effective in combating label noise than linking them to labeled nodes that may contain noisy labels.
Based on this observation, we need to develop a robust GNN to tackle such challenges: (1) In practice, we cannot determine which labels are clean. How can we differentiate between clean labels and noisy labels? (2) How can we alleviate the difficulties posed by label scarcity, which hinders nodes from receiving adequate supervision from labeled neighbors? (3) Assuming we have identified the noisy labels, how can we effectively utilize these noisily labeled nodes? 

To address the above challenges, we propose a novel robust \textbf{G}raph \textbf{N}eural \textbf{N}etwork with \textbf{C}oarse- and \textbf{F}ine-\textbf{G}rained \textbf{D}ivision for mitigating label sparsity and noise, namely GNN-CFGD. Specifically,
to address the first challenge, we model the per-node loss distribution with a Gaussian Mixture Model (GMM) \citep{permuter2006study} to dynamically divide the training nodes into cleanly labeled and noisily labeled nodes. Additionally, to reduce the confirmation bias inherent in a single network, we simultaneously train peer networks to \textit{co-decide} which nodes are clean. Further details are presented in subsection \ref{co-devide}.

Moreover, to alleviate the second challenge, we propose a clean labels oriented link to reconstruct the graph by connecting unlabeled nodes to cleanly labeled nodes. This approach differs from NRGNN \citep{dai2021nrgnn} and RTGNN \citep{qian2023robust}, which connect unlabeled nodes to labeled nodes, potentially linking them to noisy labeled nodes and inadvertently propagating noise to the unlabeled nodes. More elaboration is given in subsection \ref{reconstruction}.
Finally, to tackle the third challenge, we recognize that directly discarding noisy nodes may result in the loss of useful information about the data distribution. Therefore, to provide more refined supervision for noisy labeled nodes, we further fine-grained the identified noisy labels into a confidence set and a remaining set based on predicted confidence. Additionally, to alleviate the problem of label scarcity, we generate pseudo-labels for unlabeled nodes based on predicted confidence, thereby providing additional supervision. Details are presented in subsection \ref{pseudo_label}. Our contributions can be summarized as follows:

\begin{itemize}
  \item We first investigate how different link strategies can lead to the propagation of misinformation, adversely affecting model performance. We find that connecting unlabeled nodes only to cleanly labeled nodes proved to be a more effective strategy for mitigating labeling noise.
  \item Based on this observation, we propose a novel GNN-CFGD model. Specifically, we simultaneously train peer GCNs to \textit{co-decide} which nodes are clean. Furthermore, to alleviate the limitations of label sparsity and noise, we augment the graph structure based on cleanly labeled nodes.
  \item We conduct comprehensive experiments to evaluate the proposed GNN-CFGD model on various benchmark datasets. The experimental findings convincingly showcase the effectiveness of GNN-CFGD. 
\end{itemize}

\section{Related work}
In this section, we present a concise review of the two research areas most relevant to our work: graph neural networks and graph neural networks with label noise.
\subsection{Graph neural network}
\label{related_GNN}
GNNs have emerged as a prominent approach due to their effectiveness in working with graph-structured data. These GNN models can be broadly categorized into two main groups: spectral-based methods and spatial-based methods.

Spectral-based methods aim to identify graph patterns in the frequency domain \citep{hammond2011wavelets}. \citep{bruna2013spectral} first extended the convolution to general graphs by using a Fourier basis, treating the filter as a set of learnable parameters and considering graph signals with multiple channels, but eigendecomposition requires \(O(n^3)\) computational complexity. To reduce the computational
complexity, \citep{defferrard2016convolutional} and \citep{kipf2016semi} made several approximations and simplifications.
\citep{defferrard2016convolutional} defined fast localized convolution on the graph based on Chebyshev polynomials. 
\citep{kipf2016semi} further simplified ChebNet via a localized first-order approximation of spectral graph convolutions. 
Recently, several methods have progressively improved upon GCN \citep{kipf2016semi}, such as AGCN\citep{li2018adaptive}, DGCN \cite{zhuang2018dual}, and S$^2$GC \citep{zhu2021simple}.

Conversely, spatial-based methods directly define graph convolution in the spatial domain as transforming and aggregating local information \citep{gallicchio2010graph, dai2018learning}. The first work towards spatial-based convolutional graph neural network (ConvGNN) is NN4G \citep{micheli2009neural}, which utilized a combinatorial neural network architecture where each layer has independent parameters to learn the mutual dependencies of the graph.
To identify a particularly effective variant of the general approach, MPNN \citep{gilmer2017neural} provided a unified model for spatial-based ConvGNNs.
GIN \citep{xu2018powerful} pointed out that earlier methods based on MPNNs cannot distinguish between distinct graph structures based on the embeddings they generate. 
To address this limitation, GIN proposed a theoretical framework for analyzing the expressive capabilities of GNNs to capture different graph structures.
GraphSAGE \citep{hamilton2017inductive} generated node representations by sampling and aggregating features from local neighborhoods.
GAT \citep{velivckovic2017graph} assigned distinct edge weights based on node features during aggregation. 
SGC \cite{wu2019simplifying} reduced complexity and computation by successively removing nonlinearities and collapsing weight matrices between consecutive layers.

Many other graph neural network models have been reviewed in recent surveys \citep{wu2020comprehensive, zhou2020graph}.
Almost all of these GNNs assume that the observed labels reflect the true node labels. However, real-world applications inevitably involve label noise.

\subsection{Graph neural networks with label noise}
While GNNs have achieved significant success with graph data, only a few studies have explored their robustness when faced with noisy labels. 
D-GNN \citep{hoang2019learning} firstly demonstrated that GNNs are susceptible to label noise, proposing a backward loss correction method for GNNs to deal with label noise.
LPM \citep{xia2021towards} generated a pseudo label for each node using Label Propagation (LP). 
NRGNN \citep{dai2021nrgnn} linked unlabeled nodes to labeled nodes based on an edge predictor and provided reliable pseudo labels for the unlabeled nodes. 
RTGNN \citep{qian2023robust} further corrected noisy labels by employing reinforcement supervision and consistency regularization to mitigate the risk of overfitting. 
ERASE \citep{chen2023erase} introduced a decoupled label propagation method, which pre-corrected noisy labels through structural denoising before training and incorporated prototype pseudo-labels to update representations with error resilience during training.
While the aforementioned methods have demonstrated good performance, we explore a new perspective to alleviate the effects of label noise and sparsity, guided by our motivating experiments.

\section{Preliminary}
In this paper, we focus on semi-supervised node classification with sparse and noisy node labels. In this section, we present the preliminary related to our work.
\subsection{Notation}
Let $\mathcal{G}=(\mathcal{V},\mathcal{E})$ denotes a graph, where $\mathcal{V}=\{v_i\}_{i=1}^N$ represents the node set with base $N$, and $\mathcal{E} \in \mathcal{V} \times \mathcal{V}$ denotes the edge set with $|\mathcal{E}|=E$. 
The feature matrix of $\mathcal{G}$ is noted as $\bm{X}=[\bm{x}_1,...,\bm{x}_N]^T$ where $\bm{x}_i \in \mathbb{R}^d$ is the feature vector of node $v_i$ with $d$-dimension. 
Furthermore, $\bm{A} \in \mathbb{R}^{N \times N}$ denotes the adjacency matrix,
where $\bm{A}_{ij}=1$ if there exists an edge between node $v_i$ and $v_j$, and $\bm{A}_{ij} = 0$ otherwise. $\bm{D} \in \mathbb{R}^{N \times N}$ is the diagonal degree matrix of $\bm{A}$, where $\bm{D}_{ii}=\sum_j\bm{A}_{ij}$. Let $\mathcal{Y}=\{y_1,...,y_m\}$ denotes the given label set, where $y_i \in \mathcal{Y}$ is the label of node $v_i$.
\subsection{Problem statement}
In semi-supervised node classification on graphs with noisy and limited node labels, we are provided with two sets of nodes: a small portion set $\mathcal{V}_L$ with labels and a set $\mathcal{V}_U$ without labels. Additionally, the observed labels in $\mathcal{V}_L$ are polluted by noise, represented as $\mathcal{Y}_N=\{y_1,...,y_l\}$. Our goal is to learn a robust GNN that predicts the true labels of the unlabeled nodes. This objective can be formally denoted as:
\begin{equation}
   \mathcal{\hat{Y}}_U=f(\mathcal{G},\mathcal{Y}_N, \theta),
   \label{eq1}
\end{equation}
where $\theta$ is the parameters of function $f$ and $\mathcal{\hat{Y}}_U$ is the predicted labels of unlabeled nodes.
\subsection{Graph neural network}
Most modern GNNs stack multiple graph convolution layers to learn node representations by a message-passing mechanism \citep{gilmer2017neural}. The $l$-th layer of the GNNs takes the following form:
\begin{equation}
\bm{m}_i^{(l-1)} = \mathrm{AGGREGATE}^{(l)}(\{\bm{h}_j^{(l-1)} | {v_j}\in \mathcal{N}({v_i})\}),
\label{eq_AGG}
\end{equation}

\begin{equation}
\bm{h}_i^{(l)} = \mathrm{UPDATE}^{(l)}(\bm{h}_i^{(l-1)}, \bm{m}_i^{(l-1)}),
\label{eq_UPDATE}
\end{equation}
where $\bm{m}_i^{(l-1)}$ and $\bm{h}_i^{(l-1)}$ are the message vector and the hidden embedding of node $v_i$ at the ($l$-1)-th layer, respectively.
The set $\mathcal{N}({v_i})$ consists of nodes that are adjacent to node $v_i$. $\bm{h}_i^{(0)} = \bm{x}_i$ if $l=0$. $\mathrm{AGGREGATE}^{(l)}(\cdot)$ and $\mathrm{UPDATE}^{(l)}(\cdot)$ are characterized by the specific model, respectively. 

\section{Methodology}
In this section, we present the proposed robust GNN model, GNN-CFGD. We begin with the overview of GNN-CFGD in subsection \ref{overview}. Subsequently, we detail the three main parts of GNN-CFGD: coarse-grained division based on GMM in subsection \ref{co-devide}, clean labels oriented link in subsection \ref{reconstruction}, and fine-grained division based on confidence in subsection \ref{pseudo_label}.

\subsection{\label{overview}Overview}
\begin{figure}
    \centering
    \includegraphics[width=\linewidth]{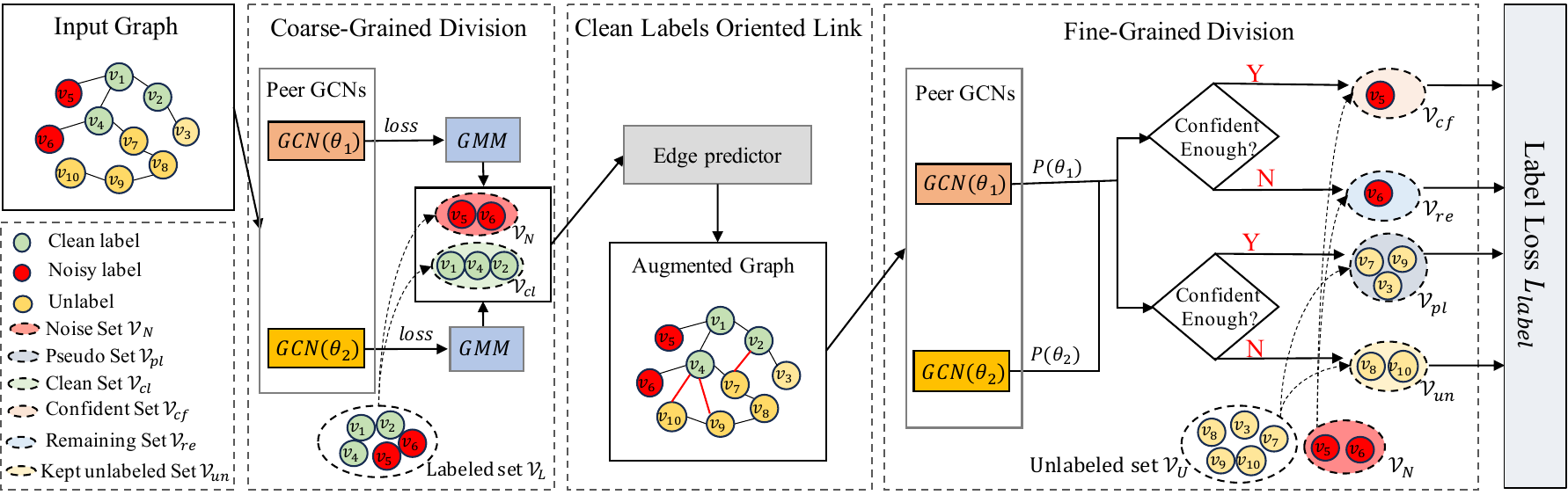}
    \caption{An illustration of the proposed model GNN-CFGD, which consists of (i) coarse-grained division based on GMM, (ii) clean labels oriented link, and (iii) fine-grained division based on confidence.}
    \label{frame}
\end{figure}

Motivated by our observations, we propose a novel robust GNN model, GNN-CFGD, designed for graphs with sparse and noisy node labels. An illustration of GNN-CFGD is presented in Fig. \ref{frame}. The model comprises three main components: Coarse-grained division based on GMM: This component incorporates a GMM along with peer GCNs, with two GCNs using different parameters. The objective is to perform a coarse-grained partition of label set $\mathcal{V}$ into clean label set $\mathcal{V}_{cl}$ and noisy label set $\mathcal{V}_N$. Clean labels oriented link: This part focuses on reconstructing the graph by linking unlabeled nodes to cleanly labeled nodes, which are derived from the co-decision of the peer GCNs. Fine-grained division based on confidence: This component aims to further refine noisy labels and unlabeled nodes. Specifically, given the augmented graph, we can divide noise label set $\mathcal{V}_N$ into a confident set $\mathcal{V}_{cf}$ and remaining nodes set $\mathcal{V}_{re}=\mathcal{V}_N-\mathcal{V}_{cf}$. Unlabeled node set $\mathcal{V}_U$ is similarly divided into a set of nodes with pseudo-labels $\mathcal{V}_{pl}$ and a set of kept unlabeled nodes $\mathcal{V}_{un}=\mathcal{V}_U-\mathcal{V}_{pl}$. Finally, to prevent overfitting, we incorporate consistency regularization based on Kullback-Liebler (KL) divergence, inspired by \citep{qian2023robust}. Each part of GNN-CFGD is detailed below.

\subsection{\label{co-devide}Coarse-grained division based on GMM}
Some existing GNNs combat label sparsity and noise by directly linking unlabeled nodes to labeled ones, which may inadvertently connect to noisy labeled nodes. However, we experimentally verify that this operation diminishes the classification performance of the model (see Fig. \ref{fig_motivate2}). An intuitive solution is to link unlabeled nodes only to cleanly labeled ones. To implement this, we developed an approach by fitting a GMM on per-node cross-entropy loss distribution, enabling us to partition the training set $\mathcal{V}_L$ into a cleanly labeled set $\mathcal{V}_{cl}$ and a noisily unlabeled set $\mathcal{V}_N$.

Recent research has demonstrated that deep models (including GNN) have the appealing phenomenon that they initially memorize clean samples and then gradually memorize noisy samples as the training epoch increases \citep{zhang2021understanding, arpit2017closer, arazo2019unsupervised}. This suggests that it is possible to differentiate clean nodes from noisy ones based on “small loss”.
Inspired by \citep{li2020dividemix}, we employ GMM with two components to fit the cross-entropy loss distribution for clean and noisy nodes. 
The probability density function (PDF) of the mixture model related to the loss $\ell$ is defined as follows:
\begin{equation}
    p(\ell)=\sum_{k=1}^2\lambda_kp(\ell|k),
    \label{eq_gmm}
\end{equation}
where $p(\ell|k)$ is the individual pdf of the $k$-th component, and $\lambda_k$ denotes the mixing coefficients for the convex combination of $p(\ell|k)$. 

For each node $v_i$, its clean probability $p_i$ is the posterior probability $p(k=0|\ell_i)$, which is formally defined as follows:
\begin{equation}
    p_i = p(k=0|l_i) = \frac{p(k)p(\ell_i|k)}{p(\ell_i)},
    \label{eq_wi}
\end{equation}
where $k=0$ is the component with a smaller mean. 

Next, for one of the peer GCNs, given a clean probability threshold $p_{th}$, the set of clean nodes can be obtained as follows:
\begin{equation}
    \mathcal{V}_{cl1} = \{v_i | p_i >p_{th}, \forall v_i \in \mathcal{V}_L\}
    \label{eq_sing_Vcl}
\end{equation}

However, experimental evidence indicates that a single network can be overly confident in its mistakes \citep{tarvainen2017mean}. To mitigate this confirmation bias, we propose a \textit{co-decide} strategy. 
As illustrated in Fig. \ref{frame}, we can obtain two distinct sets of clean labels ($\mathcal{V}_{cl1}$ and $\mathcal{V}_{cl2}$) by fitting a GMM to the outputs (i.e., the cross-entropy loss of each labeled node) from two GCNs with different parameters.
The intersection of these two clean label sets yields the clean label set for this epoch:
    \begin{equation}
        \mathcal{V}_{cl} = \mathcal{V}_{cl1} \cap \mathcal{V}_{cl2}
        \label{eq_vcl}
    \end{equation}
    
\subsection{\label{reconstruction}Clean labels oriented link,}
To alleviate label sparsity and promote supervision propagation, \citep{dai2021nrgnn} and \citep{qian2023robust} add edges between unlabeled and labeled nodes.
However, our motivational experiments indicate that connecting unlabeled to clean labeled nodes is more effective for improving the classification performance of GCN. Therefore, in this section, we augment the input graph based on the clean label set obtained through the aforementioned division.

Inspired by \citep{dai2021nrgnn}, we introduce an edge predictor with encoder-decoder \citep{kipf2016variational} to predict links. A GCN \citep{kipf2016semi} serves as the encoder to learn node representations as follows:
\begin{equation}
    \bm{Z}=\mathrm{GCN}(\bm{A},\bm{X}).
    \label{eq_encoder}
\end{equation}

We then employ a cosine-similarity decoder for link prediction. Specifically, the edge weight $w_{ij}$ between nodes $v_i$ and $v_j$ is calculated as follows:
\begin{equation}
    \bm{W}_{ij} = \mathrm{Relu}(\frac{\bm{z}^i \cdot \bm{z}^j}{\lVert \bm{z}^i \rVert \lVert \bm{z}^j \rVert}),
\end{equation}
where $\bm{z}^i$ and $\bm{z}^j$ are the representations of node $v_i$ and $v_j$, respectively. The closer $\bm{z}^i$ and $\bm{z}^j$ are, the higher $\bm{W}_{ij}$ becomes.
 
In most graphs, positive samples (i.e., node pairs with edges) tend to be much smaller than negative samples (i.e., node pairs without edges). To reduce computational complexity, we train the edge predictor by using the adjacency matrix reconstruction loss based on negative sampling \citep{mikolov2013distributed}.
The loss function is formally defined as follows:
\begin{equation}
    \mathcal{L}_{rec}=\sum_{v_i \in V}(\sum_{v_j \in \mathcal{N}(v_i)}(\bm{W}_{ij}-\bm{A}_{ij})^2 + N_{neg} \cdot \mathbb{E}_{v_n\sim P_n(v_i)}(\bm{W}_{in}-\bm{A}_{in})^2 ),
    \label{eq_lrec}
\end{equation}
where $N_{neg}$ denotes the number of negative samples for each node, and $\mathbb{E}_{v_n\sim P_n(v_i)}$ is the distribution of negative samples. With the GCN-based edge predictor trained with Eq. (\ref{eq_lrec}), we could obtain the adjacent matrix $\hat{\bm{A}}$ of the augmented graph as follows:
\begin{equation}
    \hat{\bm{A}}_{ij} = \left\{
    \begin{array}{ll}
    1 & \mathrm{if} \, \bm{A}_{ij}=1;  \\
    \bm{W}_{ij} & \mathrm{if} \, \bm{A}_{ij} =0, j \in \mathcal{V}_{cl}, \mathrm{and} \, \bm{W}_{ij} > \tau;  \\
    0 & \mathrm{otherwise}.
    \end{array}
    \right.
    \label{eq_augment}
\end{equation}
where $\mathcal{V}_{cl}$ is a clean label set and $\tau$ is a threshold to filter out edges with small weights.

\subsection{\label{pseudo_label}Fine-grained division based on confidence}
We can access the clean label set $\mathcal{V}_{cl}$ and the noisy label set $\mathcal{V}_N$ after the coarse-grained division. While $\mathcal{V}_{cl}$ can be directly used for supervision, we need to consider how to handle $\mathcal{V}_N$.
Discarding noisy labeled nodes directly risks losing valuable information and results in sparser training data, which can adversely affect the model's learning capacity. Conversely, directly using noisy label nodes may result in overfitting of the model. 

To this end, we adopt a compromise: we further divide the identified noisy labels set $\mathcal{V}_{N}$ into a confidence set $\mathcal{V}_{cf}$ and a remaining set $\mathcal{V}_{re}$ based on predicted confidence and consistency. This approach allows for finer-grained supervision from the noisily labeled nodes while providing additional supervision from the unlabeled nodes, ultimately enhancing the robustness of GNNs.

Based on the augmented graph structure and node features, we can learn two predicted distributions using the peer GCNs as follows:
\begin{equation}
    \bm{P}_{\theta_1} = \mathrm{GCN}(\hat{\bm{A}}, \bm{X}, \theta_1),
    \label{eq_predict1}
\end{equation}
\begin{equation}
    \bm{P}_{\theta_2} = \mathrm{GCN}(\hat{\bm{A}}, \bm{X}, \theta_2).
    \label{eq_predict2}
\end{equation}

We relabel the nodes in $\mathcal{V}_{N}$ based on the predictions (i.e. $\bm{P}_{\theta_1}$ and $\bm{P}_{\theta_2}$) of two GCNs. The pseudo label of node $v_i$ in $\mathcal{V}_N$ is defined as follows:
\begin{equation}
\begin{aligned}
    & \hat{z}^i = \mathrm{arg} \max\limits_c \bm{P}_{\theta_1}^{i,c} \\
    & \text{s.t.} \quad \begin{cases}
         &  \mathrm{arg} \max\limits_c \bm{P}_{\theta_1}^{i,c} =  \mathrm{arg} \max\limits_c \bm{P}_{\theta_2}^{i,c} \\
        & \mathrm{arg} \max\limits_c \bm{P}_{\theta_1}^{i,c} \neq y_i \\
        & \sqrt{\bm{P}_{\theta_1}^{i,\hat{z}^i} \cdot \bm{P}_{\theta_2}^{i,\hat{z}^i}} > th_{pse1}
    \end{cases}
\end{aligned}
\label{eq_vcf}
\end{equation}
where $y^i$ is the observed label of node $v_i$ and $th_{pse1}$ is the confidence threshold. The nodes that satisfy Eq. (\ref{eq_vcf}) form a confidence set $\mathcal{V}_{cf}$. Nodes in $\mathcal{V}_{cf}$ are used for supervision, while the nodes in $\mathcal{V}_{re} = \mathcal{V}_N - \mathcal{V}_{cf}$ are down-weight in the supervision process.

In practical applications, labeled nodes are often scarce. To effectively increase the amount of available labeled data and alleviate the issue of label sparsity, we generate pseudo-labels for the unlabeled nodes. The pseudo label of node $v_i$ in the unlabeled nodes $\mathcal{V}_{U}$ is defined as follows:
\begin{equation}
\begin{aligned}
    & \tilde{z}^i = \mathrm{arg} \max\limits_c \bm{P}_{\theta_1}^{i,c} \\
    & \text{s.t.} \quad \begin{cases}
         &  \mathrm{arg} \max\limits_c \bm{P}_{\theta_1}^{i,c} =  \mathrm{arg} \max\limits_c \bm{P}_{\theta_2}^{i,c} \\
        & \sqrt{\bm{P}_{\theta_1}^{i,\tilde{z}^i} \cdot \bm{P}_{\theta_2}^{i,\tilde{z}^i}} > th_{pse2}
    \end{cases}
\end{aligned}
\label{eq_vpl}
\end{equation}

where $th_{pse2}$ is the confidence threshold. The nodes that satisfy Eq. (\ref{eq_vpl}) can form the pseudo-label set  $\mathcal{V}_{pl}$, Consequently, the retained unlabeled node set is given by $\mathcal{V}_{un} = \mathcal{V}_U -\mathcal{V}_{pl}$.

\subsection{Model learning}
Firstly, we defined the loss metric for a node $v_i$ in the peer GCNs using cross-entropy as follows:
\begin{equation}
    \mathcal{L}_{pe} = -(y_i log\bm{P}_{\theta_1}^i + y_i log\bm{P}_{\theta_2}^i) = -y_i log(\bm{P}_{\theta_1}^i \cdot \bm{P}_{\theta_2}^i).
\end{equation}

To date, we have partitioned the training set into five subsets, i.e., clean set $\mathcal{V}_{cl}$, confident set $\mathcal{V}_{cf}$, the remaining set $\mathcal{V}_{re}$, the pseudo-label set $\mathcal{V}_{pl}$, and the kept unlabeled set $\mathcal{V}_{un}$. The overall loss for these labeled nodes can be expressed as follows:
\begin{equation}
    \mathcal{L}_{label} =\frac{1}{|\mathcal{V}_L|} \sum_{v_i \in \mathcal{V}_L} \omega(i)\hat{y}_i log(\bm{P}_{\theta_1}^i \cdot \bm{P}_{\theta_2}^i),
    \label{eq_label}
\end{equation}
where 
\begin{equation}
    \left\{
       \begin{array}{ll}
        \omega(i)=1, \,  \hat{y}_i =y_i & \mathrm{if} \, v_i \in \mathcal{V}_{cl};  \\
        \omega(i)=1, \,  \hat{y}_i =\hat{z}_i & \mathrm{if} \, v_i \in \mathcal{V}_{cf};  \\
        \omega(i)=\beta, \,  \hat{y}_i =y_i & \mathrm{if} \, v_i \in \mathcal{V}_{re}; \\
        \omega(i)=1, \, \hat{y}_i =\tilde{z}_i & \mathrm{if} \, v_i \in \mathcal{V}_{pl}
        \end{array}
    \right. 
    \label{eq_weight}
\end{equation}

Additionally, to prevent overfitting, inspired by \citep{qian2023robust}, we add consistency regularization for labeled node $v_i$ based on KL divergence as follows:
\begin{equation}
    \mathcal{L}^i_{reg} = \mathcal{L}^i_{inter} + \mathcal{L}^i_{intra},
    \label{eq_loss_reg}
\end{equation}
where
\begin{equation}
     \mathcal{L}^i_{inter} = D_{KL}(\mathcal{P}^i_{\theta_1} || \mathcal{P}^i_{\theta_2}) + D_{KL}(\mathcal{P}^i_{\theta_2} || \mathcal{P}^i_{\theta_1}),
\end{equation}
\begin{equation}
     \mathcal{L}^i_{intra} = \sum_j \frac{\hat{\bm A}_{ij}}{\sum_k\hat{\bm{A}}_{ik}}(D_{KL}(\mathcal{P}^j_{\theta_1} || \mathcal{P}^i_{\theta_1}) + D_{KL}(\mathcal{P}^j_{\theta_2} || \mathcal{P}^i_{\theta_2})),
\end{equation}

Finally, by combining the graph reconstruction loss (Eq. (\ref{eq_lrec})), the total training loss is expressed as follows:
\begin{equation}
    \mathcal{L}_{total} = \mathcal{L}_{label} + \alpha \mathcal{L}_{rec} + \lambda \mathcal{L}_{reg} 
    \label{eq_total_loss}
\end{equation}
where $\alpha$ and $\lambda$ are balance parameter.

Based on the previous description, the training algorithm for GNN-CFGD is outlined in Algorithm \ref{algorithm}. 
Specifically, the algorithm begins by initializing all the parameters of GNN-CFGD. In lines 6-11, GNN-CFGD performs a coarse-grained partition of the observed labels into a clean label set and a noisy label set. Lines 12-13 detail the clean labels oriented linking process, which aims to reconstruct the graph. In lines 14-17, the noisy label set is further divided into a confidence set and a remaining set, while the unlabeled nodes are similarly categorized into a pseudo-label set and a retained unlabeled node set. Finally, in line 22, the parameters of the peer GCNs are updated by minimizing the total loss specified in line 21. Note that both GCNs can be used for inference; in our experiments, we utilize the first one.

\begin{algorithm}
    \caption{Model training for GNN-CFGD.}\label{algorithm}
    \begin{algorithmic}[1]
    \STATE \parbox[t]{\linewidth}{\textbf{Input:} adjacency matrix  $\bm{A}$, feature matrix $\bm{X}$, labels $\mathcal{Y}_L$, $\theta_1$ and $\theta_2$, clean probability threshold $p_{th}$, number of negative samples $N_{neg}$, balance parameter $\alpha$ and $\lambda$, weight threshold $\tau$, confidence threshold $th_{pse1}$ and $th_{pse2}$, total epochs $T$.}
    \STATE initialize $\hat{\bm{A}}= \bm{A}$ 
    \STATE \textit{// warm up the model for a few epochs using the standard cross-entropy loss} 
    \STATE $\theta_1, \theta_2= \text{WarmUp}(\bm{A}, \bm{X}, \mathcal{Y}_L, \theta_1, \theta_2)$ 
    \STATE $ \textbf{for } i < T \textbf{ do} $ 
    \STATE \hspace{0.5cm} \textit{// lines 7-8: calculate per-node cross-entropy loss using peer GCNs}
    \STATE \hspace{0.5cm} $ \ell1 = \text{peerGCN1}(\hat{\bm{A}},\bm{X}, \theta_1)$ 
    \STATE \hspace{0.5cm} $ \ell2 = \text{peerGCN2}(\hat{\bm{A}},\bm{X}, \theta_2)$ 
    \STATE \hspace{0.5cm} \textit{// lines 10-11: model per-node loss using GMM to obtain clean probability}
    \STATE \hspace{0.5cm} $P_1 = \text{GMM}( \ell1)$ 
    \STATE \hspace{0.5cm} $P_2 = \text{GMM}( \ell2)$
    \STATE \hspace{0.5cm} \textit{// lines 13-15: coarse-grain observed labels into clean and noisy label sets.}
    \STATE \hspace{0.5cm} $ \mathcal{V}_{cl1} =\{v_i | p_i > p_{th}, \forall p_i\in P_1\} $
    \STATE \hspace{0.5cm} $ \mathcal{V}_{cl2} =\{v_i | p_i > p_{th}, \forall p_i \in P_2\} $
    \STATE \hspace{0.5cm} $ \mathcal{V}_{cl} = \mathcal{V}_{cl1} \cap \mathcal{V}_{cl2}, \mathcal{V}_N=\mathcal{V}_L-\mathcal{V}_{cl}$
    \STATE \hspace{0.5cm} \textit{// lines 17-20: clean labels oriented link}
    \STATE \hspace{0.5cm} $ \bm{Z}=\mathrm{GCN}(\bm{A},\bm{X})$
    \STATE \hspace{0.5cm} $ \bm{W}_{ij} = \mathrm{Relu}(\frac{\bm{z}^i \cdot \bm{z}^j}{\lVert \bm{z}^i \rVert \lVert \bm{z}^j \rVert}) $
    \STATE \hspace{0.5cm} \textit{// reconstruct graph structure with Eq. (\ref{eq_augment})}
    \STATE \hspace{0.5cm} $ \hat{\bm{A}} = \text{Update}(\bm{A}, \bm{W}, \tau) $ 
    \STATE \hspace{0.5cm} \textit{// lines 22-23: learn two predicted distributions using peer GCNs}
    \STATE \hspace{0.5cm} $ \bm{P}_{\theta_1} = \mathrm{peerGCN1}(\hat{\bm{A}}, \bm{X}, \theta_1) $
    \STATE \hspace{0.5cm} $\bm{P}_{\theta_2} = \mathrm{peerGCN2}(\hat{\bm{A}}, \bm{X}, \theta_2) $
    \STATE \hspace{0.5cm} \textit{//calculate pseudo label for nodes in $\mathcal{V}_N$ with Eq. (\ref{eq_vcf}) }
    \STATE \hspace{0.5cm} $ \hat{z}^i = \text{PseudoLabel}(\bm{P}_{\theta_1}, \bm{P}_{\theta_2}, \mathcal{Y}_N,  th_{pse1}) $ 
    \STATE \hspace{0.5cm} \textit{//calculate pseudo label for nodes in $\mathcal{V}_U$ with Eq. (\ref{eq_vpl}) }
    \STATE \hspace{0.5cm} $ \tilde{z}^i = \text{PseudoLabel}(\bm{P}_{\theta_1}, \bm{P}_{\theta_2}, th_{pse2}) $ 
    \STATE \hspace{0.5cm}  $\mathcal{L}_{total} = \mathcal{L}_{label} + \alpha\mathcal{L}_{rec} + \lambda\mathcal{L}_{reg}$  
    \STATE \hspace{0.5cm} \text{update the parameters $\theta_1$, $\theta_2$ by minimizing $\mathcal{L}_{total}$ }
    \STATE $\textbf{end}$
    \STATE $\textbf{return } \hat{\bm{A}}, \theta_1, \theta_2$
    \end{algorithmic}
\end{algorithm}

\section{Experiment}
In this section, we conduct a comprehensive evaluation to assess the effectiveness of the proposed GNN-CFGD model. First, we compare the performance of GNN-CFGD against several state-of-the-art methods in the semi-supervised node classification task on graphs with two types of noise. Additionally, we analyze the robustness of GNN-CFGD on graphs in the context of label sparsity.
Next, we perform an ablation study to validate the significance of each component within the GNN-CFGD model. Finally, we present an analysis of hyperparameter sensitivity and visualize the proportion of linked nodes with noisy nodes.
\subsection{Setting up}
\subsubsection{Datasets}
\label{subsubsec_dataset}
We conduct experiments on four commonly used benchmark datasets: Cora-ML \citep{gasteiger2018predict}, Citeseer \citep{sen2008collective}, Pubmed \citep{sen2008collective}, Coauthor CS \citep{shchur2018pitfalls}. Table \ref{table_data} provides a summary of the statistical information about these datasets. We follow the previous study, utilizing 5\% of labeled nodes per class for training, 15\% for validation, and reserving the rest for testing. In line with \citep{patrini2017making, yu2019does}, we introduce two types of label noise to the training set: (1) Uniform Noise: the labels are flipped to other classes with a certain probability $p$ in a uniform manner; (2) Pair Noise: The labels have a probability of $p$ of being flipped to their corresponding pair class. For each type of label noise, we conduct evaluations with varying values $p$ in $\{20\%, 30\%, 40\%\}$.

\begin{table}[htbp]
	\centering
         \resizebox{0.7\linewidth}{!}{
	\begin{tabular}{lrrrc}
		\hline
		Dataset     & \multicolumn{1}{c}{\#Nodes} & \multicolumn{1}{c}{\#Edges} & \multicolumn{1}{c}{\#Features} & \#Classes \\ \hline
		Cora-ML     & 2810                       & 7981                       & 2879                          & 7        \\
		CiteSeer    & 3327                       & 4732                       & 3703                          & 6        \\
		PubMed      & 19717                      & 44338                      & 500                           & 3        \\
		Coauthor CS & 18333                      & 81894                      & 6805                          & 15       \\ \hline
	\end{tabular}
         }
	\caption{The statistics of the datasets}
	\label{table_data}
\end{table}
\subsubsection{Baselines}
To demonstrate the robustness of the proposed GNN-CFGD method, we compare it with two categories of baselines: two classical GNN models (GCN \citep{kipf2016semi}, SGC \citep{wu2019simplifying} )
and four methods to deal with label noise (Co-teaching+ \citep{yu2019does}, NRGNN \citep{dai2021nrgnn}, RTGNN \citep{qian2023robust}, ERASE \citep{chen2023erase}).
The details are given as follows:

1) GCN: It employs an efficient layer-wise propagation rule, which is derived from a first-order approximation of spectral graph convolutions.

2) SGC: It alleviates the excessive complexity of GCNs by iteratively eliminating nonlinearities between GCN layers and consolidating weights into a single matrix.

3) Co-teaching+: It bridges the “Update by Disagreement” strategy with the original Co-teaching. Its primary contribution lies in simultaneously maintaining two networks that identify data with prediction disagreements.

4) NRGNN: It employs an edge predictor to identify missing links that connect unlabeled nodes with labeled ones, along with a pseudo-label miner to enhance the label set.

5) RTGNN: It is designed to enhance robust graph learning by correcting noisy labels and further generating pseudo-labels for unlabeled nodes

6) ERASE: It pre-corrects noisy labels through structural denoising before training, and during training, it combines prototype pseudo-labels to update representations with error resilience.

\subsubsection{Implementation details}
For the GCN and SGC models, we use the corresponding Pytorch Geometric library implementations \citep{fey2019fast}. 
For the Co-teaching+, NRGNN, RTGNN, and ERASE methods, we utilize the source codes provided by the authors and adhere to the settings outlined in their original papers, with careful tuning.
For our peer GCNs, we use Adam \citep{kingma2014adam} optimizer and adopt a 2-layer GCN \citep{kipf2016semi} with 128 hidden units as the backbone. The edge predictor is implemented as a GCN with 64 hidden units, with $N_{neg}=50$ for negative sampling, and $\tau=0.1$ fixed across all datasets.
We set the confidence thresholds $th_{pse1}=th_{pse2}$ to values in $\{0.7, 0.8, 0.9, 0.95\}$, and $p_{th}$ to values in $\{0.4, 0.5, 0.6, 0.7\}$. Additionally, we vary the reconstruction loss coefficient $\alpha$ from $0$ to 1, and maintain a fixed factor $\beta = 0.1$ of noisy nodes.
Finally, the proposed GNN-CFGD model is trained for 200 epochs with a weight decay of 5e-4. When reporting quantitative measurements, the tests are repeated 10 times, and the average results are presented. 
\subsection{Evaluation on label noise setting}
In this section, we evaluate the robustness of GNN-CFGD in semi-supervised node classification under two types of noise and varying noise rates. The results are presented in Table \ref{table_noise}. Based on these evaluation results, we draw the following observations:

\begin{itemize}
    \item Compared with other baselines, the GNN-CFGD shows outstanding performance across all datasets under almost all cases with different types and rates of label noise. Significantly, as the ratio of flipped labels rises—resulting in expected performance declines for baselines—our model consistently upholds superiority in making accurate predictions.
    
    \item GNN-CFGD demonstrates a significant performance improvement over the baseline GCN model. Specifically, on the PubMed dataset, as the uniform noise rate increases from 20\% to 40\%, the performance of GCN degrades by 9.03\%. In contrast, GNN-CFGD experiences only a 4.14\% reduction in performance, illustrating its robustness against this type of noise. Similarly, when the pair noise rate rises from 20\% to 40\%, GCN suffers a performance decline of 14.1\%, while GNN-CFGD's performance declines by just 6.48\%. Similar findings are observed across other datasets. These results highlight the superior ability of GNN-CFGD to maintain accuracy in the presence of label noise.
    \item Our performance improvement compared to other methods that address label noise demonstrates the effectiveness of augmentation based on clean labels. Specifically, although NRGNN and RTGNN—both of which link unlabeled nodes to clean labeled nodes—ranked second in handling uniform and pair noise on the Cora-ML and CiteSeer datasets, GNN-CFGD consistently emerged as the top performer in both noise scenarios. This highlights GNN-CFGD’s remarkable ability to leverage clean labels, reinforcing its robustness in noisy labels.
\end{itemize}
\begin{table*}[t]
	\centering
	\caption{Quantitative results $(\%\pm\sigma)$ on semi-supervised node classification task with varying label noise rates. The best-performing models are highlighted in bold, while the runners-up are underlined.}
     \resizebox{\linewidth}{!}{
	\begin{tabular}{l|l|ccc|ccc}
		\hline
		\multicolumn{1}{l|}{\multirow{2}{*}{Dataset}} & \multicolumn{1}{l|}{\multirow{2}{*}{Model}} & \multicolumn{3}{c|}{Uniform Noise}         & \multicolumn{3}{c}{Pair Noise}          \\ \cline{3-8}
		                                              &                                             & 20\%          & 30\%          & 40\%          & 20\%        & 30\%         & 40\%        \\ \hline
		\multirow{7}{*}{Cora-ML}                      & GCN                                         & 70.71$\pm$3.50   & 65.45$\pm$4.50   & 62.30$\pm$5.49    & 71.56$\pm$3.44 & 62.06$\pm$4.33  & 56.85$\pm$5.80  \\ 
		                                              & SGC                                         & 71.65$\pm$3.80   & 65.78$\pm$3.70   & 64.88$\pm$5.10   & 73.60$\pm$4.10 & 63.87$\pm$3.20  & 58.75$\pm$5.00  \\
		                                              & Co-teaching+                                & 70.09$\pm$2.99  & 70.80$\pm$3.20   & 59.85$\pm$2.31   & 70.62$\pm$2.35 & 62.55$\pm$2.18  & 59.79$\pm$3.89  \\
		                                              & NRGNN                                       & 74.94$\pm$1.65   & 73.52$\pm$2.37   & \underline{71.45$\pm$1.96}   & 76.86$\pm$2.49 & \underline{74.42$\pm$3.03}  & \underline{68.9$\pm$1.96}   \\
		                                              & RTGNN                                       & \underline{77.69$\pm$2.02}   & \underline{73.59$\pm$4.45}   & 66.68±7.12   & \underline{80.90$\pm$1.02} & 72.00$\pm$5.21 & 63.23$\pm$3.42  \\
		                                              & ERASE                                       & 75.43±2.80   & 74.56$\pm$3.08         & 69.48$\pm$3.26   & 78.47$\pm$1.66 & 71.37$\pm$3.08         & 66.83$\pm$3.41  \\
		                                              & GNN-CFGD                                    & \textbf{79.74$\pm$1.72}   & \textbf{76.88$\pm$2.54}   & \textbf{72.03$\pm$3.28}   & \textbf{81.11$\pm$1.44} & \textbf{75.01$\pm$3.59}  & \textbf{69.80$\pm$3.55} \\ \hline
		\multirow{7}{*}{CiteSeer}                     & GCN                                         & 58.53$\pm$5.05   & 50.86$\pm$7.90   & 58.69±3.75   & 59.37$\pm$5.07 & 52.40$\pm$5.21	 & 47.71$\pm$5.68  \\
		                                              & SGC                                         & 56.52$\pm$5.71   & 51.74$\pm$3.48   & 48.09$\pm$5.57   & 58.02$\pm$5.45 & 50.74$\pm$5.06  & 47.36$\pm$5.45  \\
		                                              & Co-teaching+                                & 57.46$\pm$2.49   & 54.28$\pm$2.03   & 50.22$\pm$4.54   & 59.92$\pm$1.32 & 53.46$\pm$3.25  & 53.25$\pm$3.72  \\
		                                              & NRGNN                                       & 63.59$\pm$2.27   & 60.88$\pm$4.45   & 58.56$\pm$5.08   & 65.78$\pm$3.05 & 63.85$\pm$2.87  & 61.13$\pm$3.50  \\
		                                              & RTGNN                                       & 67.97$\pm$2.15   & 65.18$\pm$3.58   & \underline{66.02$\pm$4.70}   & 69.86$\pm$1.29 & 66.83$\pm$3.91  & \underline{65.29$\pm$5.02}  \\
		                                              & ERASE                                       & \underline{72.45$\pm$1.21}   & \underline{67.89$\pm$2.49}   & 63.55$\pm$5.06   & \underline{72.13$\pm$2.62} & \underline{67.39$\pm$1.91}  & 62.87$\pm$3.25  \\
		                                              & GNN-CFGD                                    & \textbf{72.80$\pm$1.53}   & \textbf{69.51$\pm$2.00}   & \textbf{67.80$\pm$2.90}     & \textbf{72.89$\pm$1.18} & \textbf{71.35$\pm$2.08}  & \textbf{68.35$\pm$4.90}   \\ \hline
		\multirow{7}{*}{PubMed}                       & GCN                                         & 80.00$\pm$1.95   & 75.52$\pm$3.51   & 70.97$\pm$4.23   & 76.96$\pm$2.65 & 70.36$\pm$4.55  & 62.86$\pm$4.37  \\
		                                              & SGC                                         & 76.92$\pm$2.09   & 73.05$\pm$3.66   & 69.04$\pm$2.81   & 74.29$\pm$1.95 & 68.10$\pm$4.76  & 61.81$\pm$3.71  \\
		                                              & Co-teaching+                                & \underline{82.48$\pm$0.95}   & \underline{80.55$\pm$0.92}   & 75.19$\pm$2.38   & 81.51$\pm$0.77 & \underline{79.60$\pm$1.30}   & \underline{72.91$\pm$4.31}  \\
		                                              & NRGNN                                       & 82.25$\pm$0.57   & 80.59$\pm$0.78   & 75.49$\pm$2.36   & \underline{80.27$\pm$0.7}5 & 75.91$\pm$1.71  & 68.31$\pm$2.77  \\
		                                              & RTGNN                                       & 77.97$\pm$1.06   & 78.01$\pm$1.09   & 75.63$\pm$2.40   & 76.64$\pm$1.13 & 74.39$\pm$1.77  & 69.53$\pm$3.07  \\
		                                              & ERASE                                       & 80.25$\pm$0.49   &  79.27$\pm$0.77  & 77.09$\pm$1.65   & 80.21$\pm$0.63 &  78.60$\pm$0.92 & 71.57$\pm$3.10   \\
		                                              & GNN-CFGD                                    & \textbf{82.84$\pm$0.12}   & \textbf{80.95$\pm$0.64}   & \textbf{78.70$\pm$3.50}     & \textbf{82.00$\pm$1.06}  & \textbf{80.68$\pm$0.93}  & \textbf{75.52$\pm$5.20} \\ \hline
		\multirow{7}{*}{Coauthor CS}                  & GCN                                         & 87.27$\pm$2.45   & 84.64$\pm$2.78   & 81.79$\pm$4.74   & 80.59$\pm$2.76 & 74.28$\pm$3.76  & 62.40$\pm$3.76  \\
		                                              & SGC                                         & 82.93$\pm$2.05   & 78.95$\pm$4.04   & 69.18$\pm$5.10   & 77.30$\pm$2.57 & 72.27$\pm$4.03  & 62.82$\pm$5.01  \\
		                                              & Co-teaching+                                & 89.59$\pm$0.53   & 88.33$\pm$0.79   & 86.72$\pm$0.75   & 85.73$\pm$1.49 & 79.05$\pm$2.87  & 66.13$\pm$2.30  \\
		                                              & NRGNN                                       & \textbf{91.18$\pm$1.00}   & \underline{90.31$\pm$0.88}   & \underline{89.74$\pm$0.78}   & \underline{87.05$\pm$1.32} & \underline{82.11$\pm$1.57}  & \underline{75.30$\pm$3.29}  \\
		                                              & RTGNN                                       & 88.57$\pm$0.67   & 86.56$\pm$0.63   & 85.71$\pm$0.86   & 84.94$\pm$1.64 & 81.52$\pm$1.63  & 66.54$\pm$1.63  \\
		                                              & ERASE                                       & 81.24$\pm$2.51   & 82.72$\pm$2.11   & 81.36$\pm$3.65   & 79.54$\pm$3.03 & 76.43$\pm$5.22  & 75.16$\pm$2.20  \\
		                                              & GNN-CFGD                                    & \underline{90.87$\pm$0.43}   & \textbf{90.69$\pm$0.37}   & \textbf{89.85$\pm$0.73}   & \textbf{89.58$\pm$0.80} & \textbf{85.01$\pm$2.48}  & \textbf{75.90$\pm$3.20}   \\ \hline
	\end{tabular}}
	\label{table_noise}
\end{table*}

\subsection{Evaluation on label sparsity setting}
\begin{figure}[htbp]
    \centering
    \begin{minipage}[b]{0.45\linewidth}
        \includegraphics[width=\textwidth]{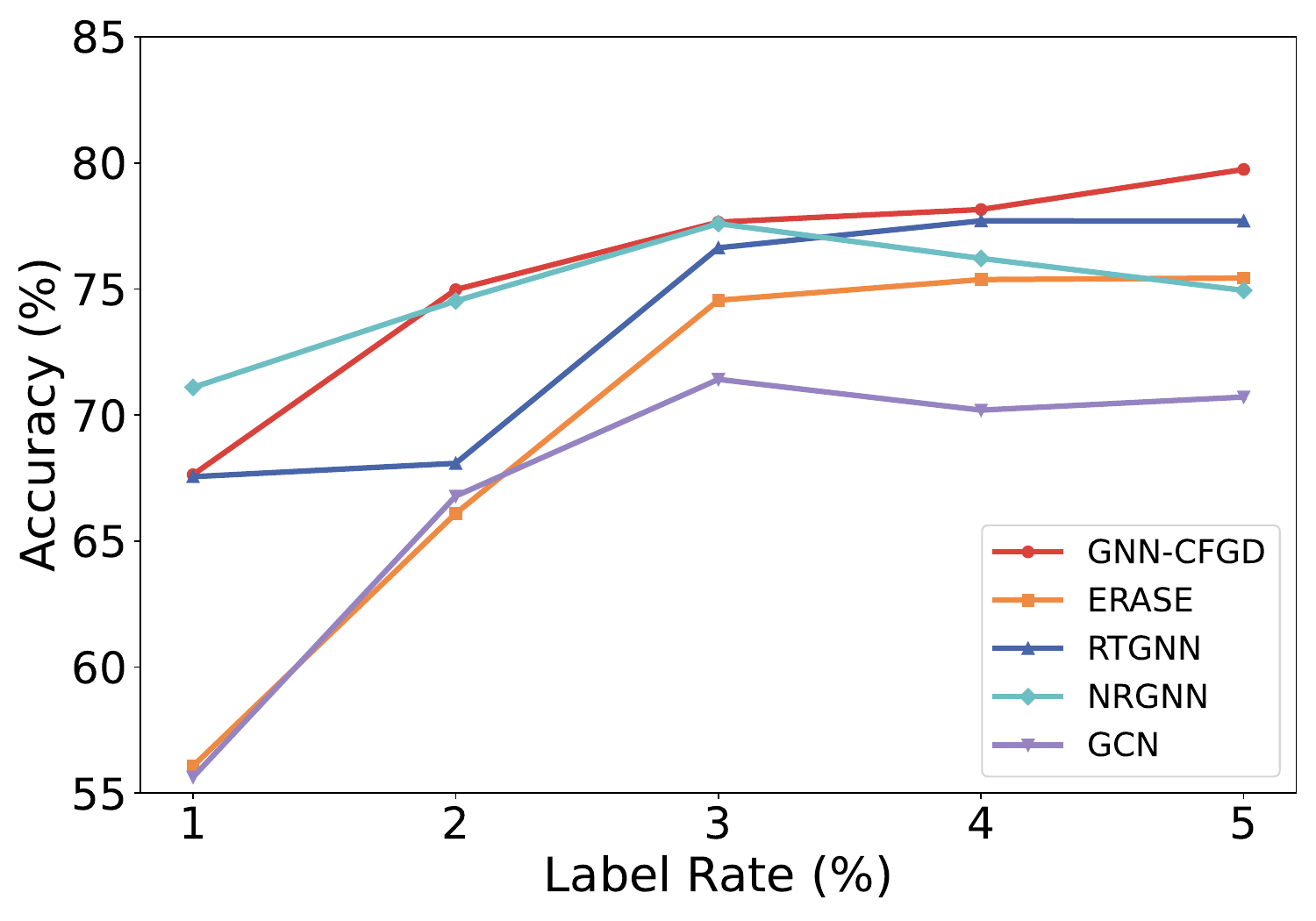}
        \centerline{\small{(a) Uniform noise}}
    \end{minipage}
    \begin{minipage}[b]{0.45\linewidth}
        \includegraphics[width=\textwidth]{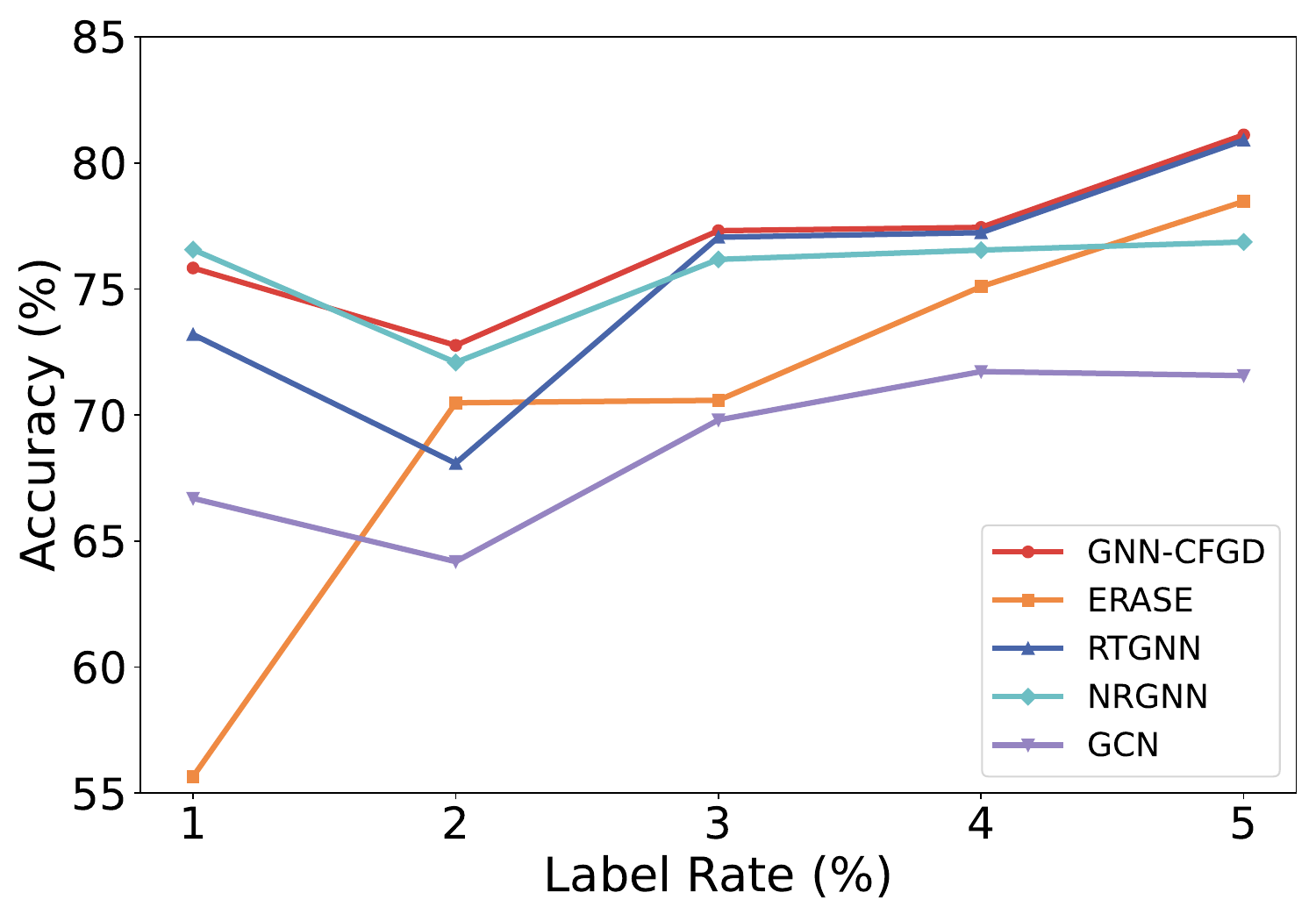}
        \centerline{\small{(b) Pair noise}}
    \end{minipage}
    \caption{Results with different label rates on Cora-ML dataset.}
    \label{fig_label_sparse_cora-ml}
\end{figure}

\begin{figure}[htbp]
    \centering
    \begin{minipage}[b]{0.45\linewidth}
        \includegraphics[width=\textwidth]{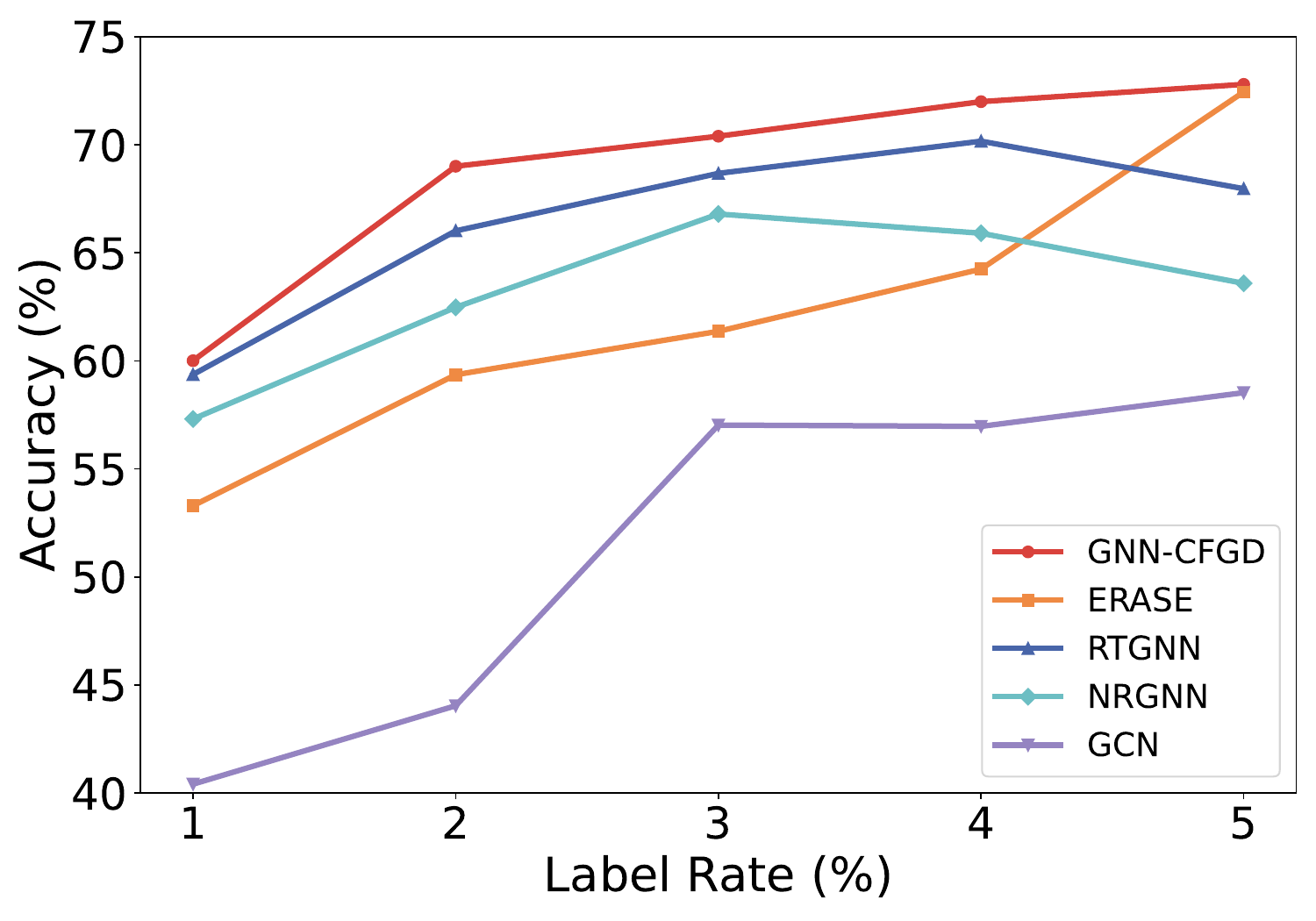}
        \centerline{\small{(a) Uniform noise}}
    \end{minipage}
    \begin{minipage}[b]{0.45\linewidth}
        \includegraphics[width=\textwidth]{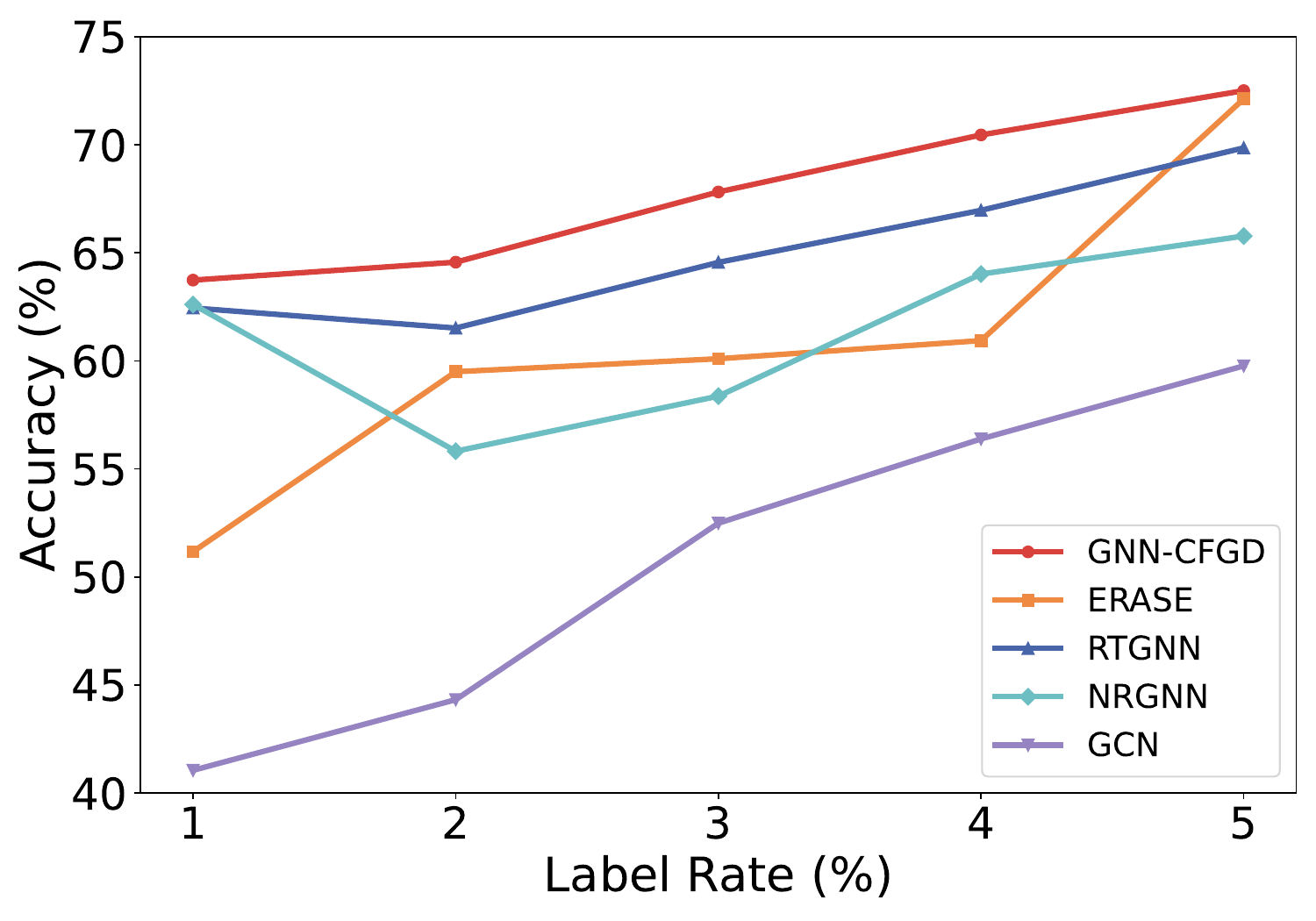}
        \centerline{\small{(b) Pair noise}}
    \end{minipage}
    \caption{Results with different label rates on CiteSeer dataset.}
    \label{fig_label_sparse_citeseer}
\end{figure}

In this subsection, we further explore the effectiveness of our proposed GNN-CFGD method under the label sparsity setting. We compare GNN-CFGD with several representative baselines (GCN, NRGNN, RTGNN, and ERASE) on Cora-ML and CiteSeer datasets with 20\% label noise rate and varying labeled training nodes (1\%, 2\%, 3\%, 4\%, 5\%). The results are presented in Fig. \ref{fig_label_sparse_cora-ml} and  Fig. \ref{fig_label_sparse_citeseer}.

As we can see, the proposed model significantly outperforms other baselines on the CiteSeer dataset across both label noise types. Specifically, as the label noise rate increases, GNN-CFGD consistently maintains a substantial performance gap between NRGNN and RTGNN, both of which also add edges between unlabeled nodes and labeled nodes. This demonstrates the effectiveness of our proposed graph enhancement based on clean labels in mitigating label sparsity.
Furthermore, although the performance distinction between GNN-CFGD and the leading baseline is not very pronounced on the Cora-ML dataset, GNN-CFGD remains the optimal choice. 
An interesting observation is that when the label rate increases from 1\% to 2\%, the accuracy of all methods declines, except for ERASE. This decline in accuracy may be attributed to the negative impact of the noise type, which appears to outweigh the benefits of increased label availability.
Combining the observations from Table \ref{table_noise}, we can conclude that GNN-CFGD can withstand label noise and mitigate label sparsity.

\subsection{Ablation study}
\begin{figure}[htbp]
    \centering
    \begin{subfigure}[t]{0.9\textwidth}
        \centering
        \begin{subfigure}[t]{0.45\textwidth}
            \centering
            \includegraphics[width=\linewidth]{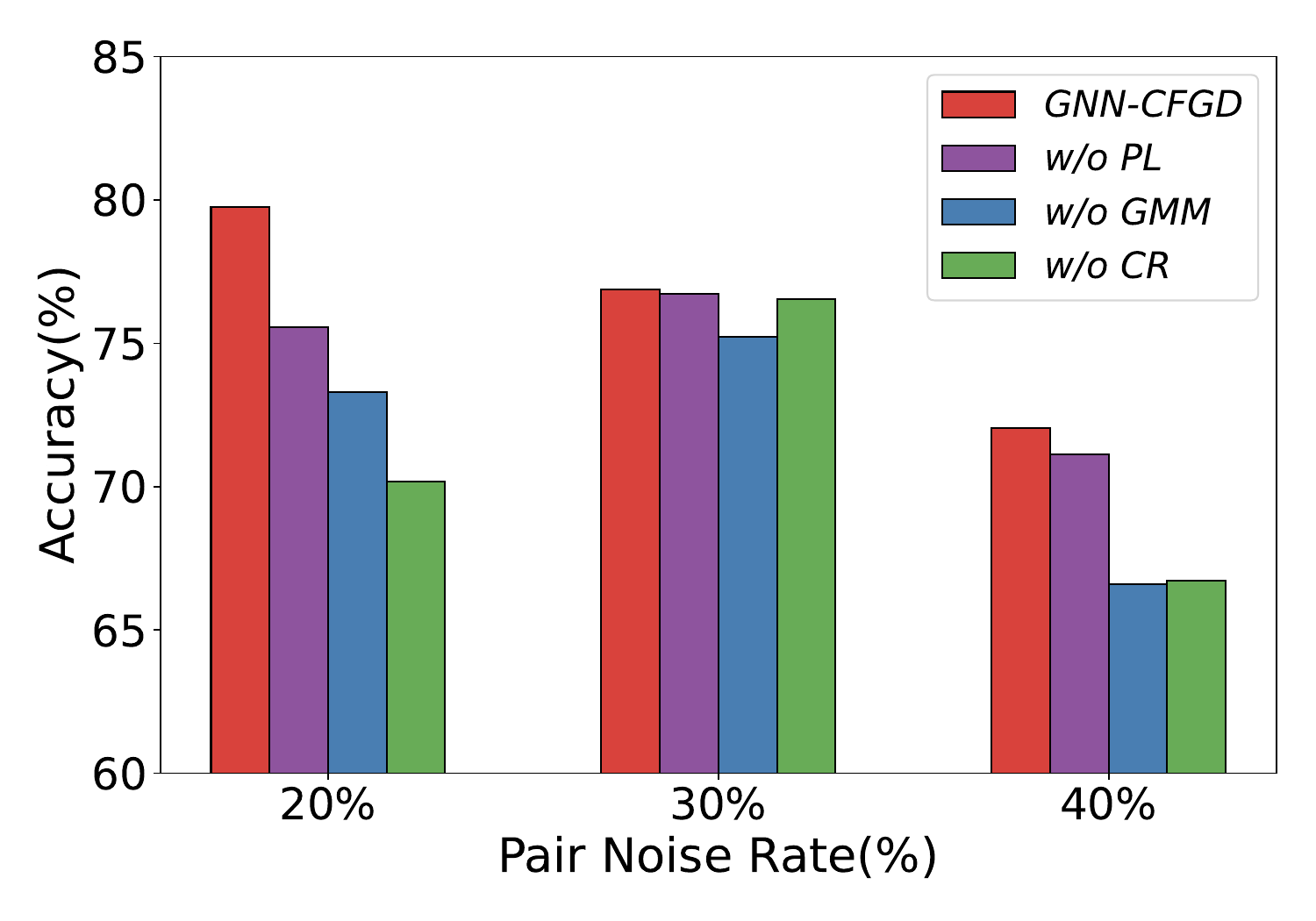}
        \end{subfigure}
        \hfill
        \begin{subfigure}[t]{0.45\textwidth}
            \centering
            \includegraphics[width=\linewidth]{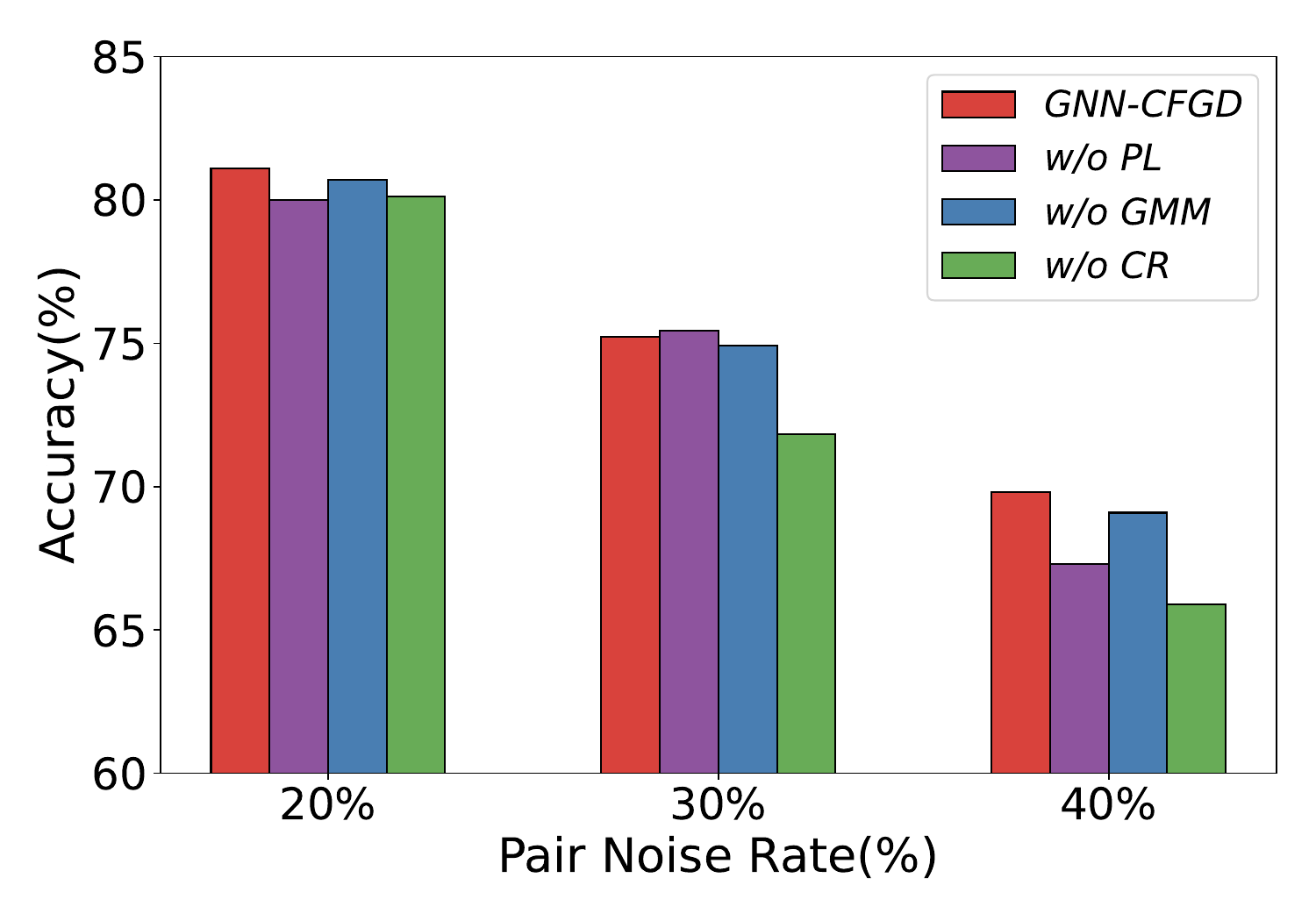}
        \end{subfigure}
        \caption{Cora-ML}
    \end{subfigure}
    \begin{subfigure}[t]{0.9\textwidth}
        \centering
        \begin{subfigure}[t]{0.45\textwidth}
            \centering
            \includegraphics[width=\linewidth]{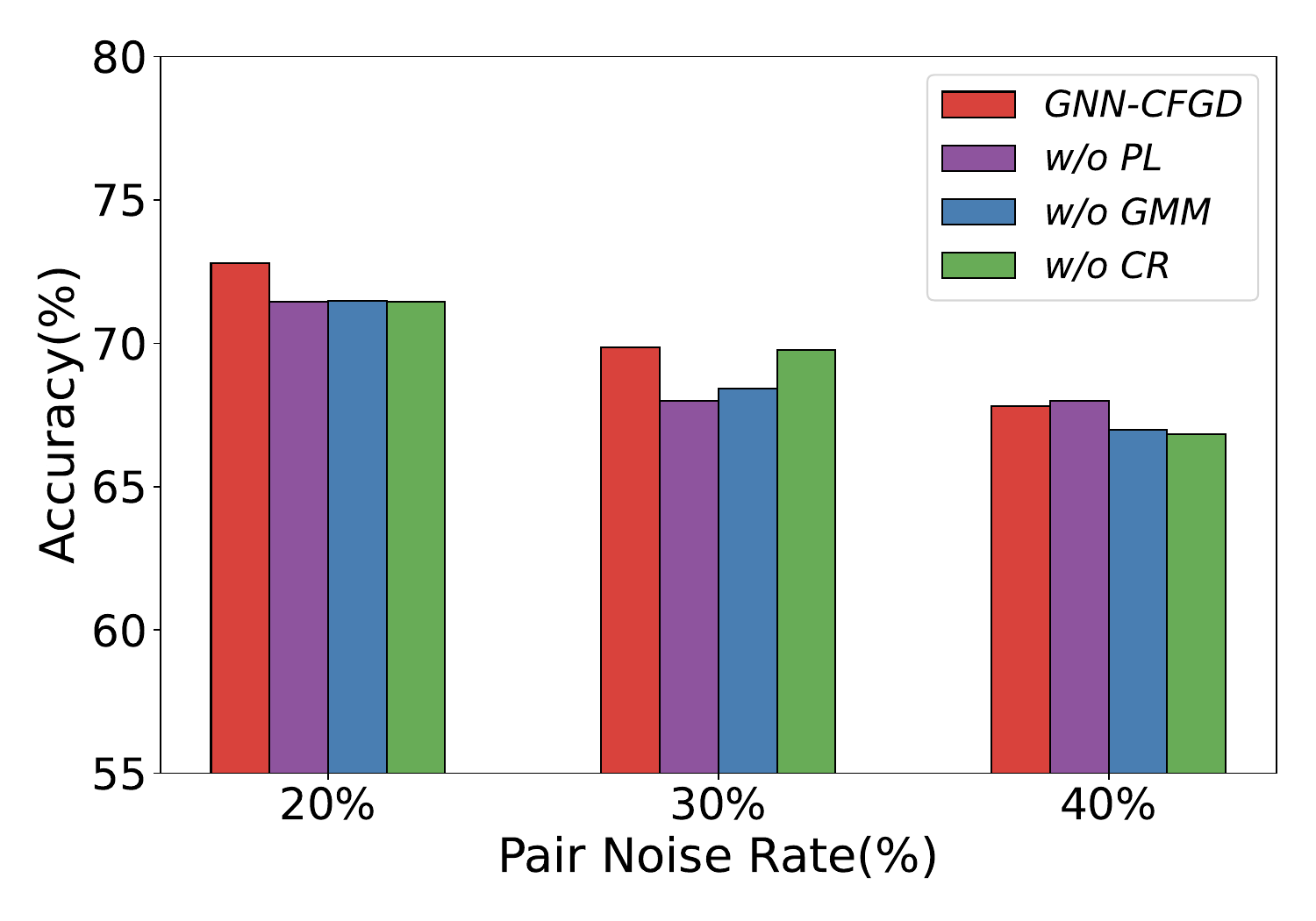}
        \end{subfigure}
        \hfill
        \begin{subfigure}[t]{0.45\textwidth}
            \centering
            \includegraphics[width=\linewidth]{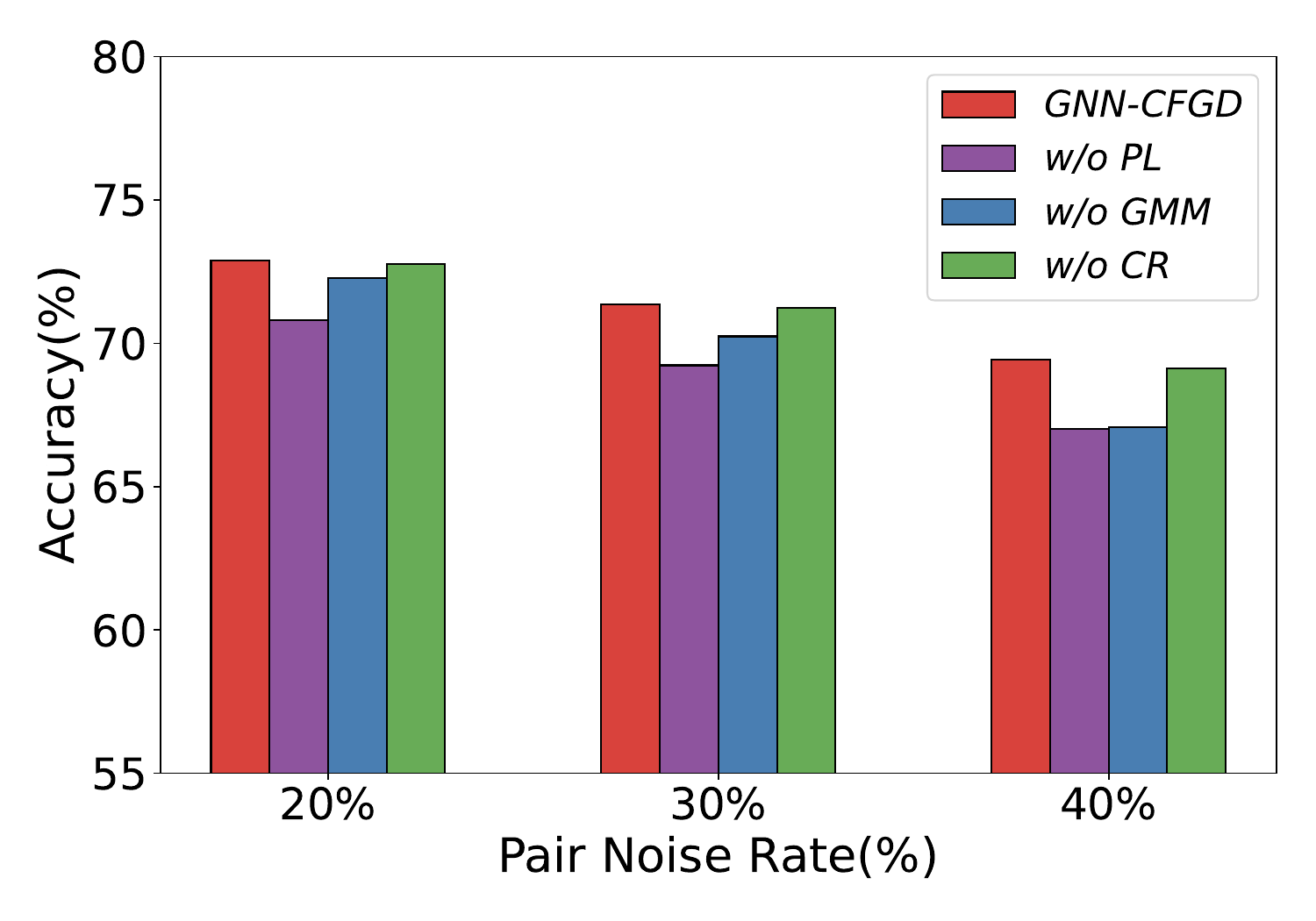}
        \end{subfigure}
        \caption{CiteSeer}
    \end{subfigure}
    \caption{Ablation study on Cora-ML and CiteSeer datasets.}
    \label{fig_ablation}
\end{figure}

In this subsection, we present the results of the ablation study examining the three variants of GNN-CFGD. Each variant is created by removing one key component from the original design of GNN-CFGD. Specifically, 1) \textit{w/o GMM} indicates that the coarse-grained division is disabled; 2) \textit{w/o PL} denotes that the fine-grained division is turned off; 3) \textit{w/o CR} refers to the variant that excludes consistency regularization for labeled nodes.
We conduct the experiment using 5\% training labels and varying the noise rates and report the results on Cora-ML and CiteSeer datasets in Fig. \ref{fig_ablation}. Our observations are as follows:
\begin{itemize}
    \item GNN-CFGD exhibits exceptional performance compared to the variant of \textit{w/o CR} across both datasets, even when subjected to two different types of noise and varying label noise rates. Notably, on the Cora-ML dataset, the proposed method demonstrates a significantly higher accuracy than \textit{w/o CR}, highlighting the importance of consistency regularization in enhancing model robustness and overall performance.
    \item The performance of the model significantly decreases when the fine-grained division is disabled except for the scenario involving uniform noise at a 40\% label noise rate on the CiteSeer dataset. This observation highlights the critical role of fine-grained division in enhancing model accuracy. In very rare cases, the reason \textit{w/o PL} win is probably we ignore graph structure information, resulting in excessive reliance on label information in confidence pseudo labeling, This is also what we need to consider in our future work.
    \item As the label noise rate increases, the performance gap between GNN-CFGD and the variant \textit{w/o GMM} widens. This phenomenon can be attributed to the fact that when the number of noisy labels is small, there are fewer connections to noisily labeled nodes when establishing links between unlabeled nodes and labeled nodes. Consequently, the impact of coarse-grained division becomes less pronounced in such scenarios. This crucial observation demonstrates that linking unlabeled nodes only to cleanly labeled nodes is beneficial for enhancing model robustness.
\end{itemize}

\subsection{Sensitivity to hyper-parameters}
\begin{figure}[htbp]
    \centering
    \begin{minipage}[b]{0.45\linewidth}
        \centering
        \includegraphics[width=\textwidth]{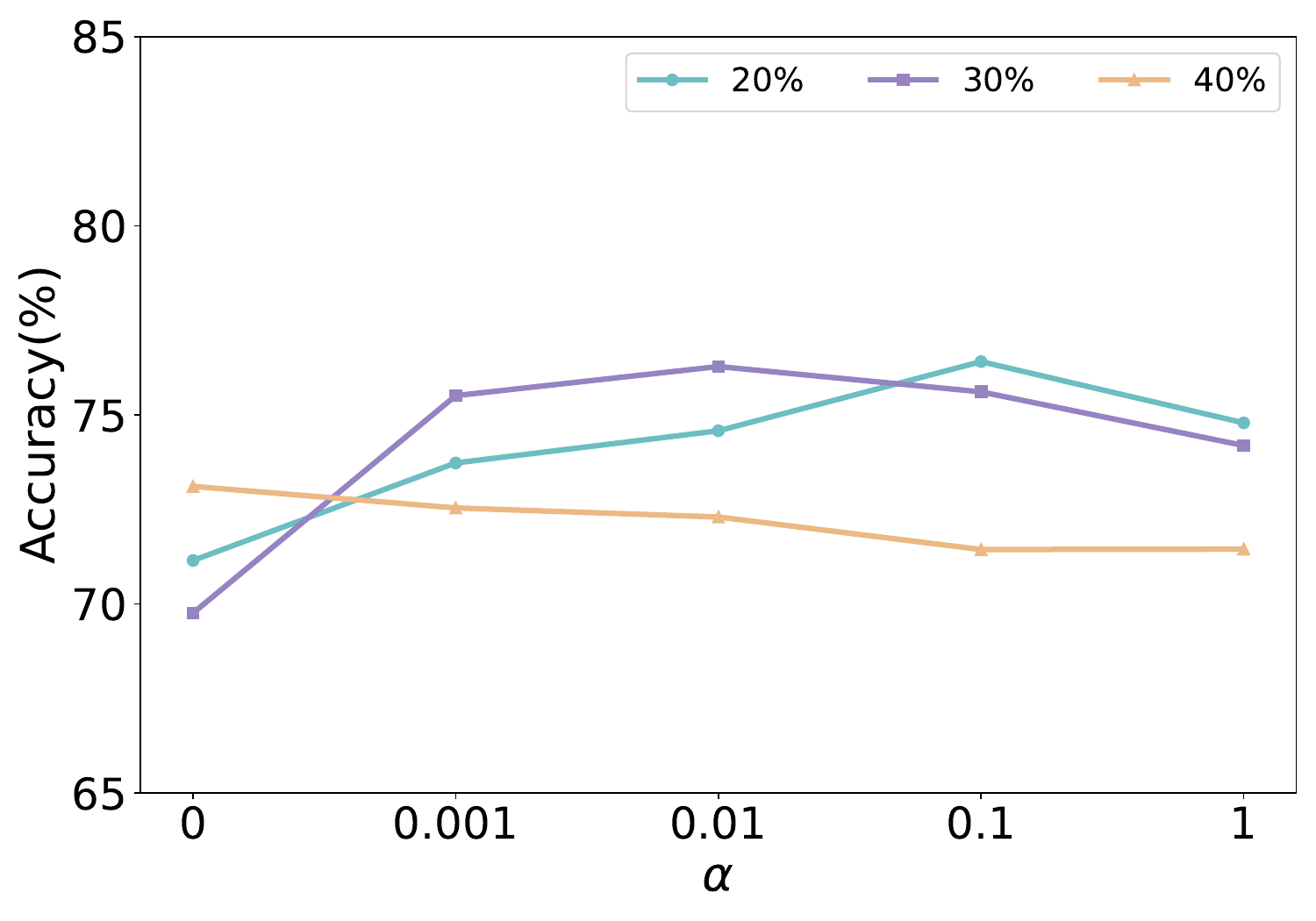}
        \centerline{\small{(a) Uniform noise}}
    \end{minipage}
    \begin{minipage}[b]{0.45\linewidth}
        \centering
        \includegraphics[width=\textwidth]{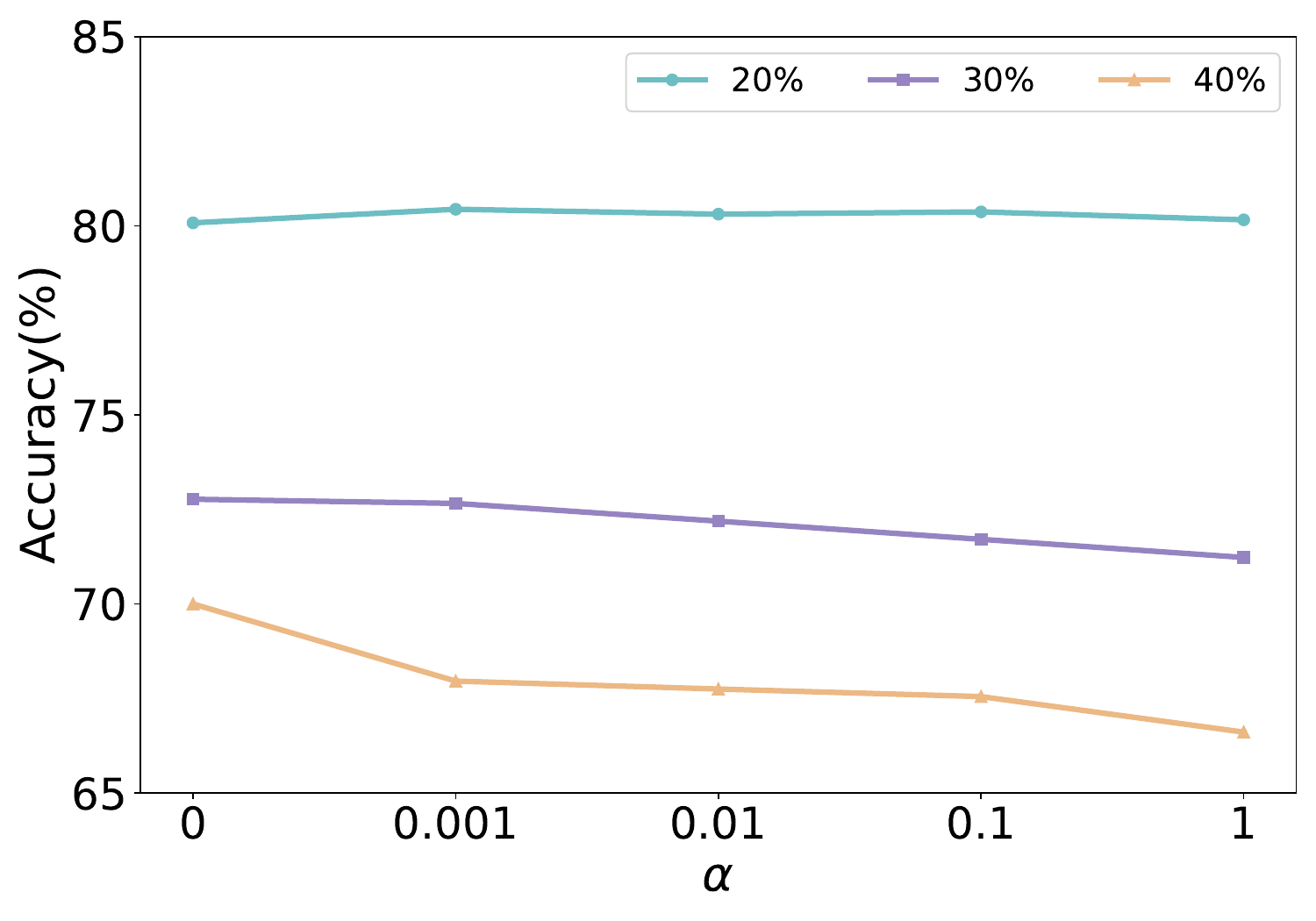}
        \centerline{\small{(b) Pair noise}}
    \end{minipage}
    \caption{Impact of hyper-parameters $\alpha$ on the Cora-ML dataset.}
    \label{fig_alpha}
\end{figure}

\begin{figure}[htbp]
    \centering
    \begin{minipage}[b]{0.45\linewidth}
        \centering
        \includegraphics[width=\textwidth]{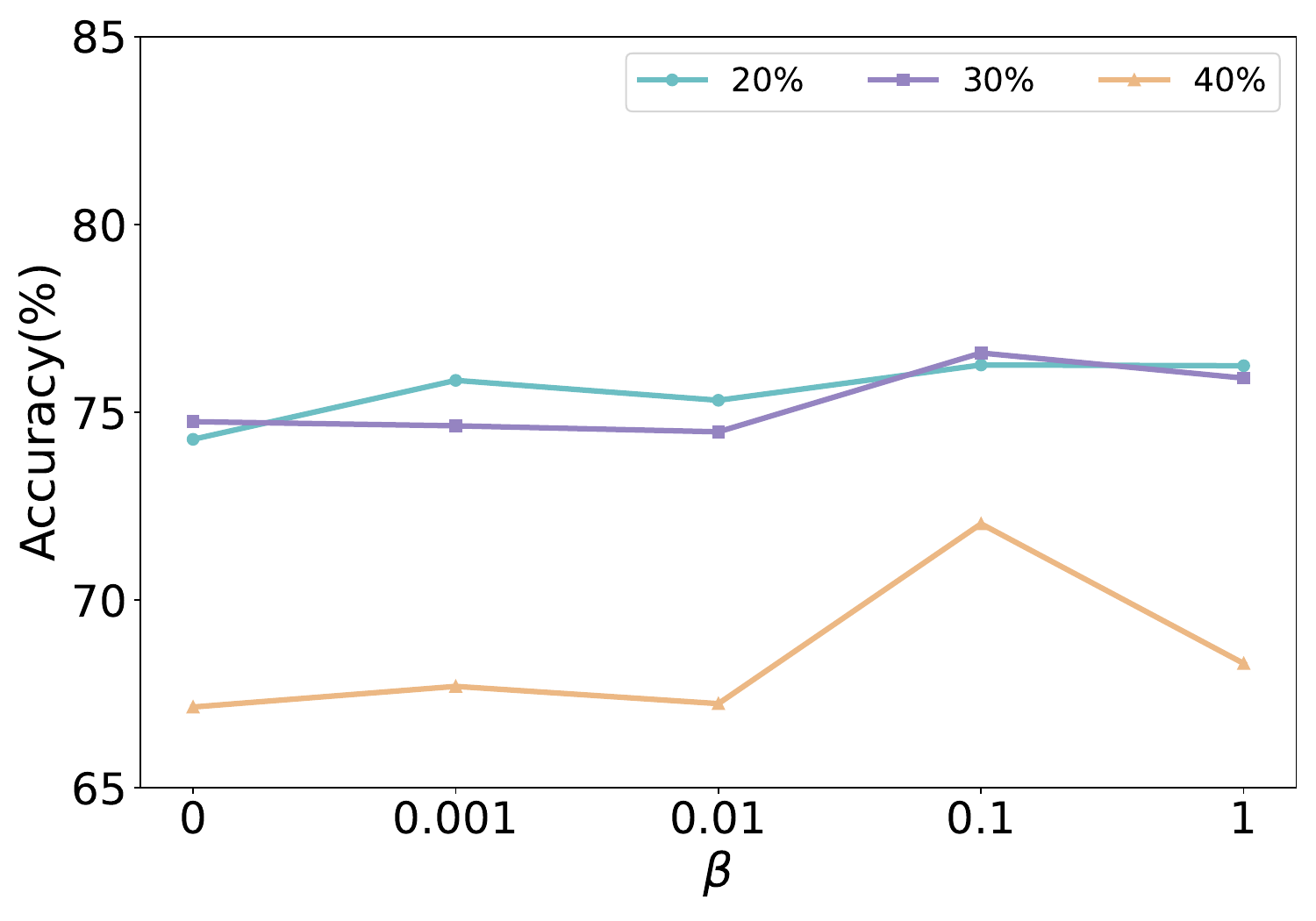}
        \centerline{\small{(a) Uniform noise}}
    \end{minipage}
    \begin{minipage}[b]{0.45\linewidth}
        \centering
        \includegraphics[width=\textwidth]{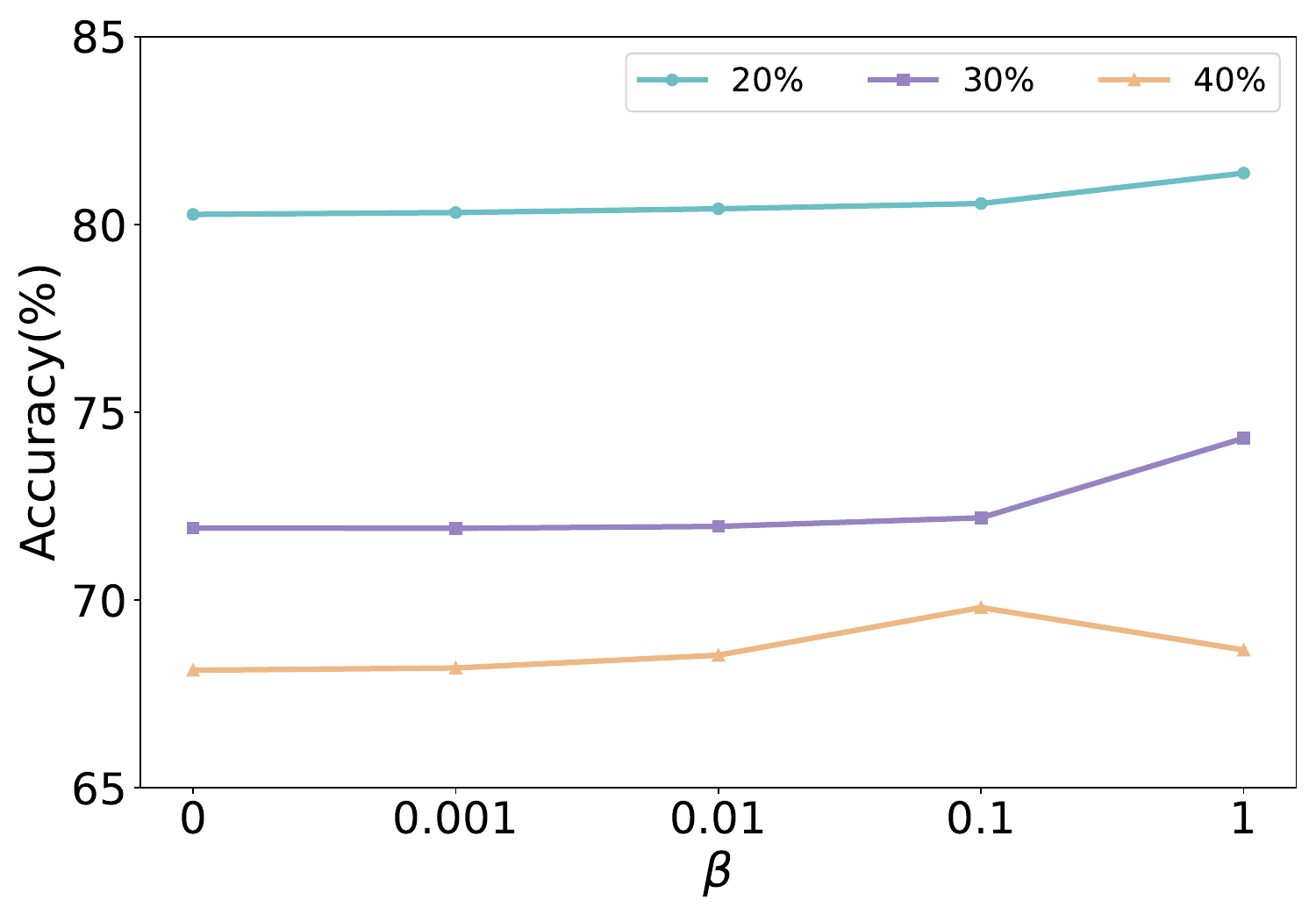}
        \centerline{\small{(b) Pair noise}}
    \end{minipage}
    \caption{Impact of hyper-parameters $\beta$ on the Cora-ML dataset.}
    \label{fig_beta}
\end{figure}

\begin{figure}[htbp]
    \centering
    \begin{minipage}[b]{0.45\linewidth}
        \centering
        \includegraphics[width=\textwidth]{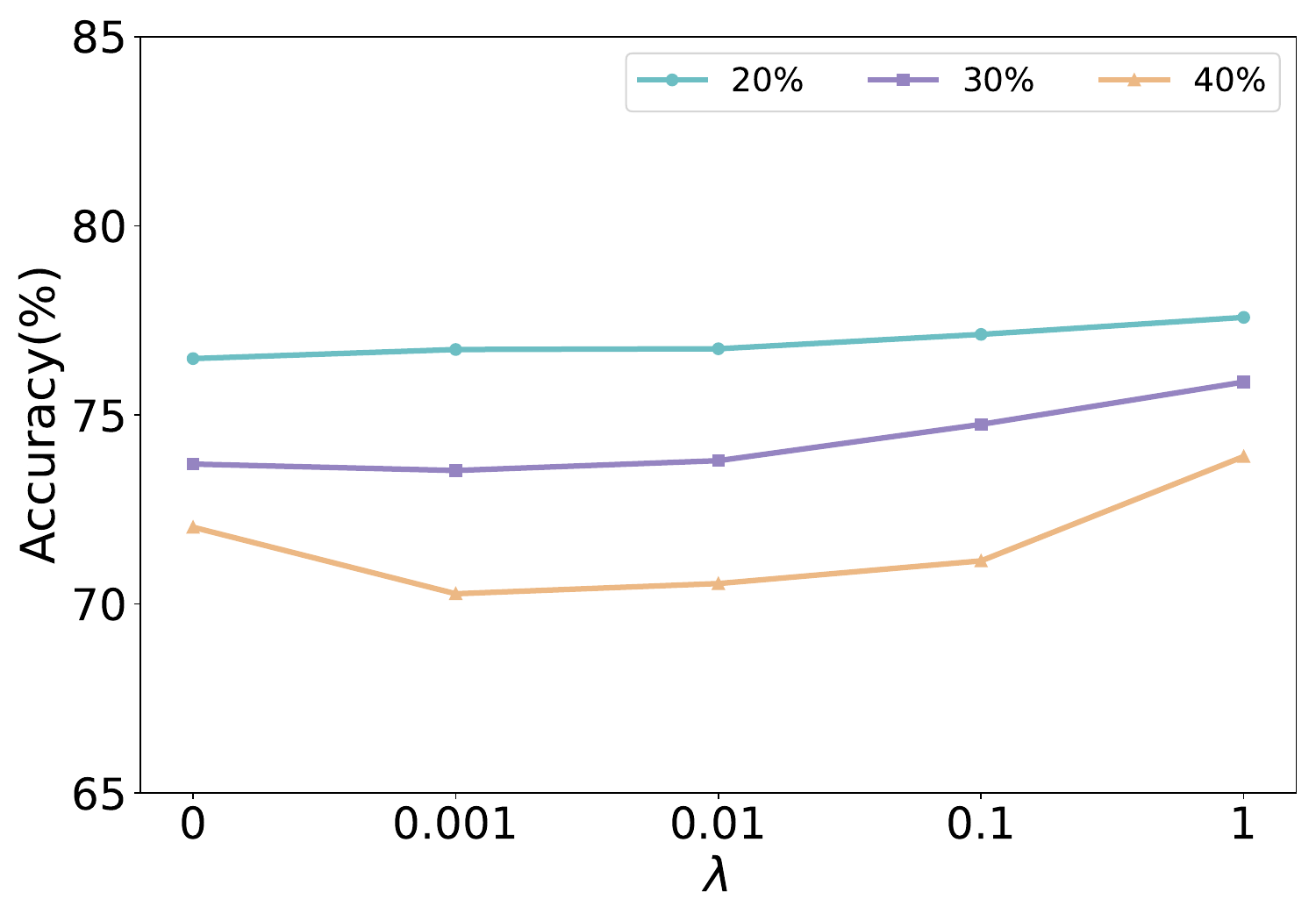}
        \centerline{\small{(a) Uniform noise}}
    \end{minipage}
    \begin{minipage}[b]{0.45\linewidth}
        \centering
        \includegraphics[width=\textwidth]{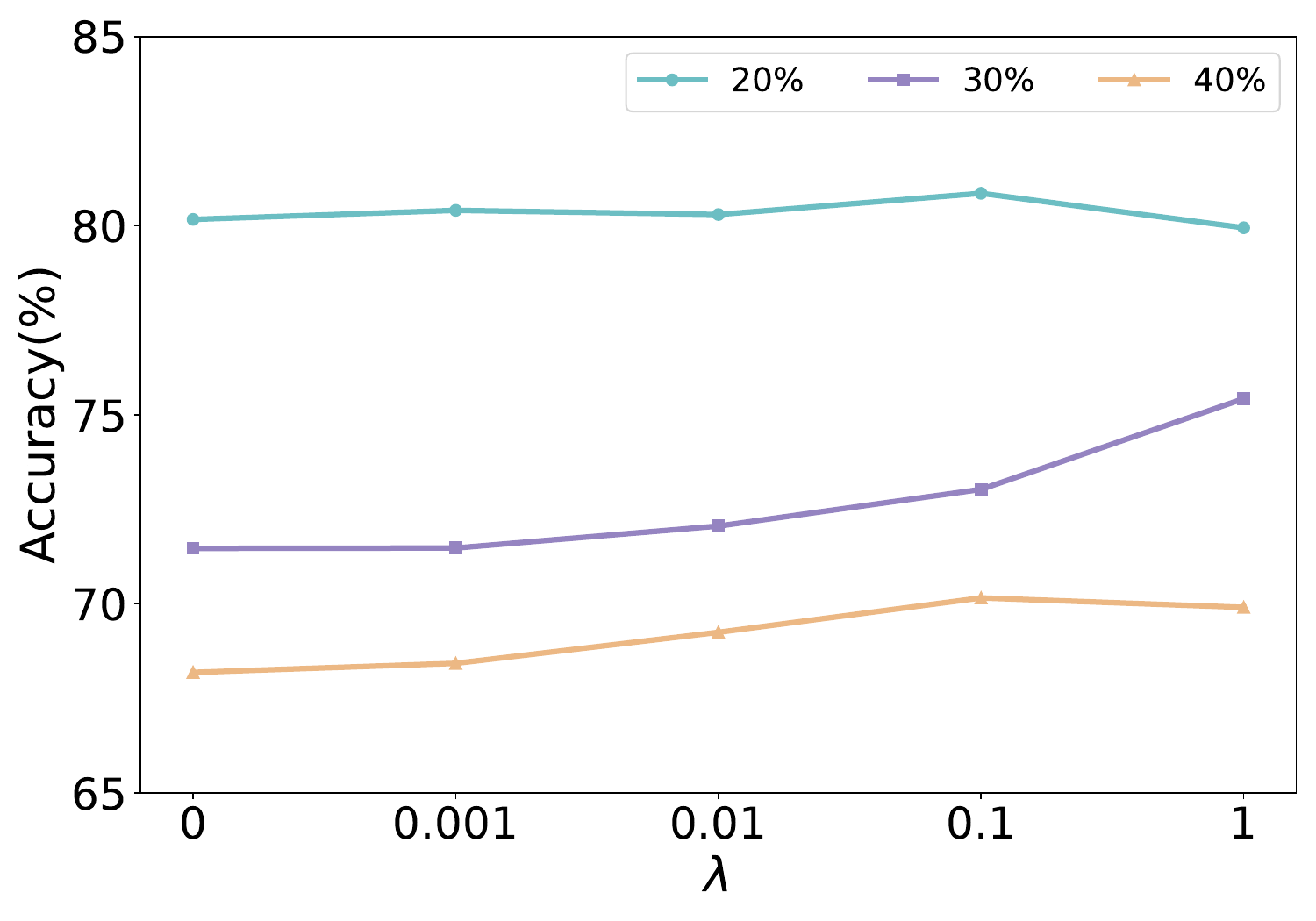}
        \centerline{\small{(b) Pair noise}}
    \end{minipage}
    \caption{Impact of hyper-parameters $\lambda$ on the Cora-ML dataset.}
    \label{fig_lambda}
\end{figure}

In this subsection, we investigate the sensitivity of key several hyper-parameters: weight coefficient $\beta$ in Eq. (\ref{eq_weight}), $\alpha$ and $\lambda$ in Eq. (\ref{eq_total_loss}). 
To conduct a comprehensive analysis, we systematically vary the values of $\beta$, $\alpha$, and $\lambda$ to assess their impact on the performance of our proposed GNN-CFGD model. 
Specifically, we explore the effects of varying $\alpha$, $\beta$, and $\lambda$ from 0 to 1 For clarity, we report the node classification accuracy of the proposed method on the Cora-ML dataset. This analysis encompasses both types of label noise and considers varying label noise rates. The results are illustrated in Figs. \ref{fig_alpha}, \ref{fig_beta}, and \ref{fig_lambda}.

We can first observe a common phenomenon: as the parameter values increase from low to high, the model's performance at a noise rate of 40\% consistently falls below the performance observed at noise rates of 20\% and 30\%. The only exception to this trend occurs in the case of uniform noise, where the parameter $\alpha$ is set to 0.
This finding indicates that a higher noise rate significantly hampers the model's ability to accurately classify nodes.
Moreover, when the noise type is set to pair noise, GNN-CFGD exhibits insensitivity to the selection of the parameters $\beta$, $\alpha$, and $\lambda$ in comparison to uniform noise. This suggests that the robustness of GNN-CFGD is less affected by variations in these hyper-parameters under pair noise conditions.
Ultimately, although the model is sensitive to parameters under uniform label noise, particularly the $\alpha$ parameter, appropriate parameter values can significantly improve performance.

\subsection{\label{visualization}Visualization the proportion of linked nodes with noisy nodes}
\begin{figure}[htbp]
    \centering
    \begin{minipage}[b]{0.4\linewidth}
        \centering
        \includegraphics[width=\textwidth]{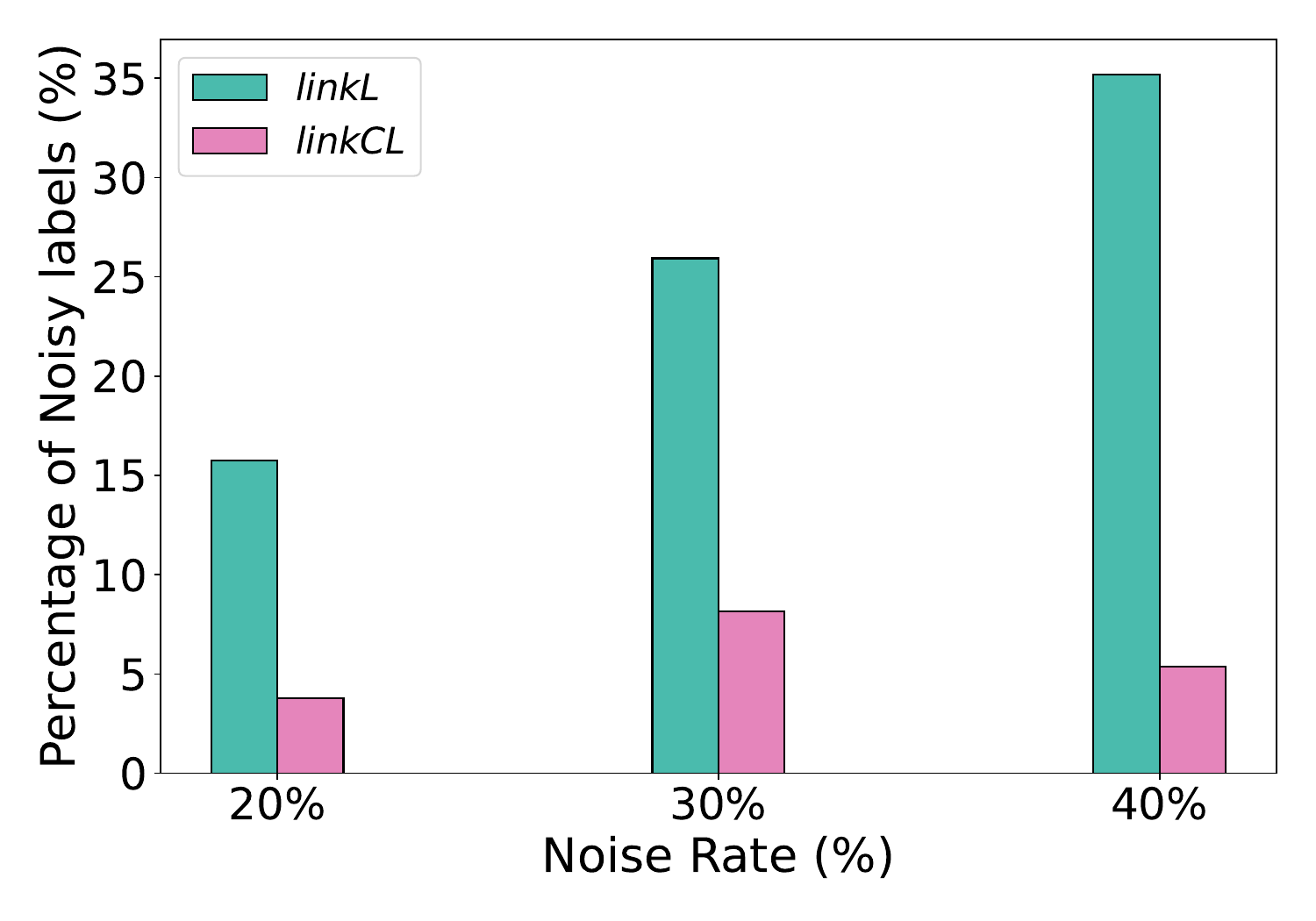}
        \centerline{\small{(a) Uniform noise}}
    \end{minipage}
    \begin{minipage}[b]{0.4\linewidth}
        \centering
        \includegraphics[width=\textwidth]{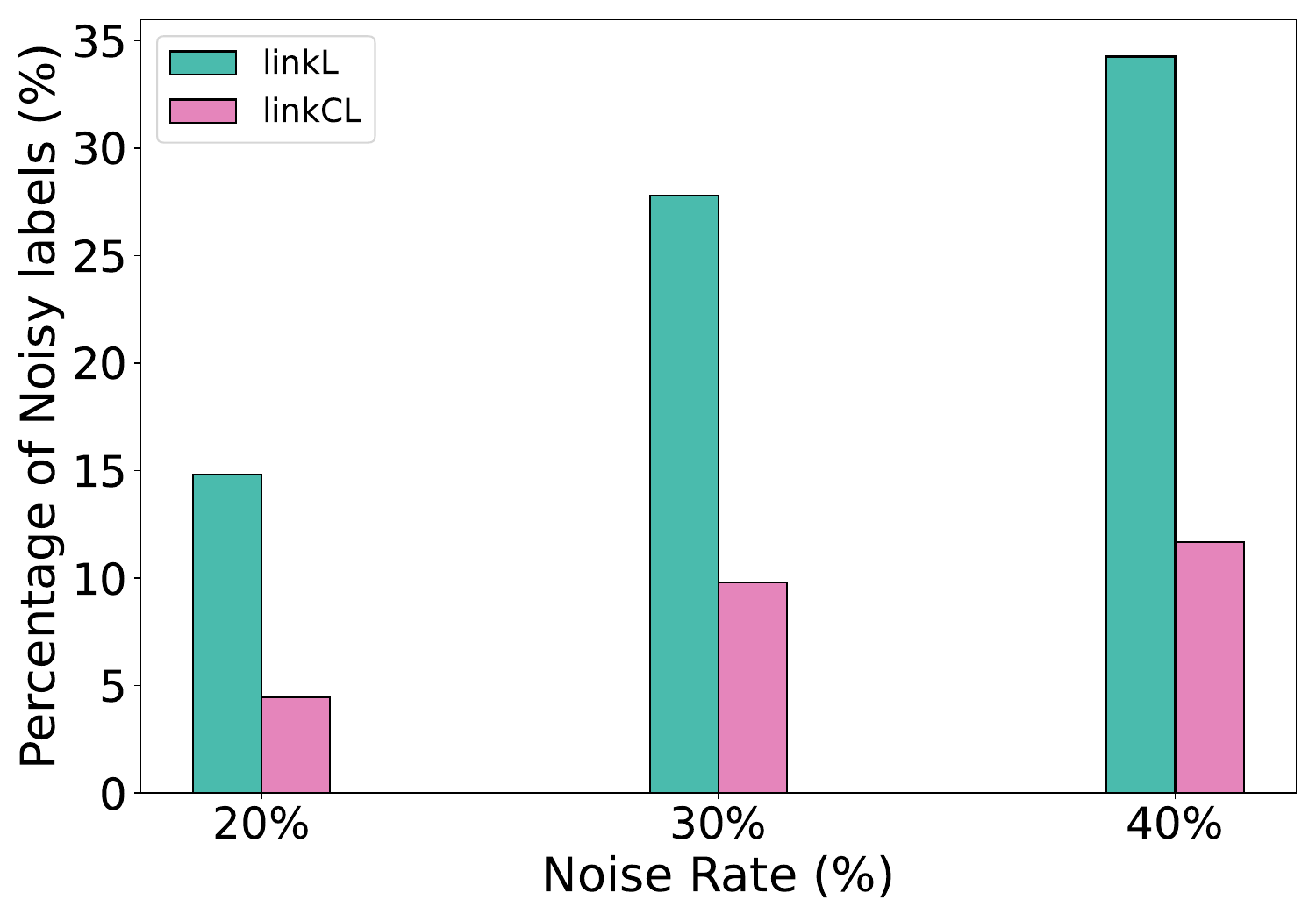}
        \centerline{\small{(b) Pair noise}}
    \end{minipage}
    \caption{Percentage of noisily labeled nodes of \textit{linkL} and \textit{linkCL} on the CiteSeer dataset.}
    \label{fig_visual}
\end{figure}

To investigate the effectiveness of GMM based coarse-grained division in identifying noisy labels, we visualize the proportion of noisy labeled nodes among those connected using two different linking strategies: \textit{linkL} and \textit{linkCL}. In our proposed method, we utilize \textit{linkCL}, while \textit{linkL} represents methods such as NRGNN \citep{dai2021nrgnn} and RTGNN \citep{qian2023robust}. To ensure consistency in the number of connected labeled nodes and the parameter settings, we adjust the linking method to either \textit{linkL} or \textit{linkCL} within the GNN-CFGD framework on the CiteSeer dataset. The experimental results are presented in Fig. \ref{fig_visual}.

We observe that the number of connected noisy nodes when using \textit{linkCL} is significantly lower than that with \textit{linkL}. This suggests that fine-grained division effectively identifies noisy labeled nodes and confirms that the proposed GNN-CFGD can mitigate the propagation of noisy labels.

\section{Conclusion}
In this paper, we propose a novel robust GNN method called GNN-CFGD, designed to tackle the challenges posed by noisy and sparse labels in semi-supervised learning tasks. GNN-CFGD is composed of three main components: 1) Coarse-grained division based on GMM, which splits train labels into clean labels and noisy labels; 2) Clean labels oriented link, which involves linking unlabeled nodes to cleanly labeled nodes to enrich the graph structure; 3) Fine-grained division based on prediction confidence, which divides the noisily labeled nodes and the unlabeled nodes into fine granularity.
The extensive experiments conducted across various datasets validate the superior performance and robustness of GNN-CFGD, demonstrating that linking unlabeled nodes to cleanly labeled nodes is significantly more effective at mitigating label noise than traditional methods.




\bibliographystyle{elsarticle-harv}
\bibliography{refs.bib}

\end{document}